\documentclass[lettersize,journal]{IEEEtran}
\usepackage{amsmath,amsfonts}
\usepackage{algorithmic}
\usepackage{algorithm}
\usepackage{array}
\usepackage[caption=false,font=normalsize,labelfont=sf,textfont=sf]{subfig}
\usepackage{textcomp}
\usepackage{stfloats}
\usepackage{url}
\usepackage{verbatim}
\usepackage{graphicx}
\usepackage{cite}

\usepackage{booktabs}
\usepackage{hyperref}
\usepackage{tikz}

\usepackage[edges]{forest}
\usepackage{subfig}
\usepackage{pifont}
\usepackage{tabularx}

\usepackage{pdfpages}

\usepackage{color}

\usetikzlibrary{shadows, arrows.meta, trees}

\newlength{\extralength}
\newlength{\fulllength}
\usetikzlibrary{shadows.blur}
\usetikzlibrary{shadows}

\newcommand{\cmark}{\ding{51}}
\newcommand{\xmark}{\ding{55}}

\begin{document}

\title{Training Machine Learning models at the Edge: A Survey}

\author{Aymen Rayane Khouas, Mohamed Reda Bouadjenek, Hakim Hacid, and Sunil Aryal
\IEEEcompsocitemizethanks{\IEEEcompsocthanksitem 
A.R. Khouas, M. R. Bouadjenek, and A. Aryal
are with the School of Information Technology, Deakin University, Waurn Ponds Campus, Geelong, VIC 3216, Australia.
H. Hacid is with the Technology Innovation Institute, UAE.\protect\\
E-mail: {\it a.khouas@deakin.edu.au} (corresponding author)}
\thanks{Manuscript received XXX YY, ZZZZ; revised XXX YY, ZZZZ.}}

\markboth{Journal of \LaTeX\ Class Files,~Vol.~14, No.~8, August~XXXX}%
{Khouas \MakeLowercase{\textit{et al.}}:}

\IEEEpubid{0000--0000/00\$00.00~\copyright~2024 IEEE}

\maketitle

\begin{abstract}
  Edge computing has gained significant traction in recent years, promising enhanced efficiency by integrating artificial intelligence capabilities at the edge. While the focus has primarily been on the deployment and inference of Machine Learning (ML) models at the edge, the training aspect remains less explored. This survey, {explores the concept of edge learning}, specifically the optimization of ML model training at the edge. The objective is to comprehensively explore diverse approaches and methodologies in {edge learning}, synthesize existing knowledge, identify challenges, and highlight future trends. Utilizing {Scopus and Web of science} advanced search, relevant literature on edge learning was identified, revealing a concentration of research efforts in distributed learning methods, particularly federated learning. This survey further provides a guideline for comparing techniques used to optimize ML for edge learning, along with an exploration of {the} different frameworks, libraries, and simulation tools available. In doing so, the paper contributes to a holistic understanding of the current landscape and future directions in the intersection of edge computing and machine learning, paving the way for informed comparisons between optimization methods and techniques designed for training on the edge.
\end{abstract}

\begin{IEEEkeywords}
Machine Learning; Edge Computing; Edge AI; Edge Learning; On-Device Training; Edge intelligence; Artificial Intelligence; IoT.
\end{IEEEkeywords}

\section{Introduction}
\label{sec:introduction}

{In recent years}, the fields of Artificial Intelligence (AI) and Machine Learning (ML) have witnessed significant growth, and have demonstrated remarkable success across various industrial applications~\cite{maslej2023artificial}. {ML's essence lies in the interplay between algorithmic models and large quantities of data, as the latter is often required to successfully train ML models}. Traditionally, datasets have been collected in cloud storage, databases, and data lakes. {These datasets are then processed} in central cloud servers to train various ML models.

Conversely, the rapid proliferation of smart devices and sensors in recent years has led to an explosion of data generation at the edge of the network. With edge devices generating vast quantities of data {closer to the source, growing concerns about privacy and security}, as well as the desire to optimize the bandwidth consumption on the increasing number of edge devices and reduce the computational load on cloud servers, have driven a paradigm shift towards edge computing. In this context, computational processes are decentralized and migrated to edge devices. This sets the stage for a novel intersection between ML and edge computing.

This shift has sparked growing interest towards edge ML. A union between machine learning and edge computing, {deploying ML models} at the edge, closer to end devices, {enabling inference or training to occur at the edge}. {Edge learning is a subset of Edge ML that involves training ML models directly at the edge}. Traditionally, ML models have relied on cloud infrastructure for training and deployment. However, this approach poses several challenges. These include high latency, significant communication overheads, and concerns around data privacy and security. {By processing data closer to its source, edge learning tackles these challenges while enabling real-time decision-making and reducing cloud resource usage}. Furthermore, {this enables innovative ML applications}, such as privacy-aware recommendation systems and smart technologies. {These applications span} multiple industries, including healthcare, manufacturing, agriculture, and space exploration.

Training ML models at the edge poses unique challenges {due to edge devices' limited computational power and memory}. Moreover, despite the abundance of data at the edge, {individual devices usually lack sufficient data to train ML models from scratch}. To address {these challenges, techniques like} federated learning, knowledge distillation, and transfer learning {have been proposed}. These methods aim to optimize ML models to fit within the constraints of edge devices, thereby rendering them suitable for training in resource-constrained environments, scenarios with low data availability, or through collaborative training across multiple edge devices that leverage their collective data.
\IEEEpubidadjcol

This survey paper aims to provide an overview on edge learning, covering its methodologies, requirements, applications, challenges, and open research directions. We explore state-of-the-art techniques for training and optimizing ML models on edge devices, highlighting their advantages. {Furthermore, we compare these approaches, providing a broad overview of their strengths and weaknesses}. We also examine the various applications that benefit from edge learning, as well as the frameworks, libraries, and simulation tools {that support and optimize it}. {Please note that a background in ML and deep learning is assumed for this survey, and readers without this knowledge may find it helpful to consult a general introduction to the field, such as the ones found in~\cite{raschka_machine_2022, lecun_deep_2015, goodfellow_deep_2016, deisenroth_mathematics_2020}.}

\subsection{{Comparison with existing surveys}}\label{sec:comparison-surveys}

There have been considerable surveys about Edge ML that attempts to define the field and present the different approaches that exist for AI in Edge. Most of these surveys focus on edge inference or on a single aspect of edge learning, such as federated learning~\cite{abreha_federated_2022, boobalan_fusion_2022} or on-device training~\cite{zhu_-device_2023, Dhar2021SurveyOnDevice}. 

As previously defined, this survey provides a comprehensive overview of training ML models on edge devices. This survey has multiple contributions that we will use to compare with the existing surveys. The contributions are presented in the following five points that we will label as "topics".
\begin{enumerate}
    \item \textbf{Explore techniques}: {We examine various techniques for optimizing the training of ML models on edge devices.}
    \item \textbf{Metrics for Edge Learning}: {We define metrics to evaluate and compare edge learning approaches, and identify requirements for edge learning in real-world scenarios.}
    \item \textbf{Compare techniques}: {We compare the different edge learning techniques based on their performance, requirements, and popularity.}
    \item \textbf{Explore Types of ML}: {We examine various types of ML, including unsupervised and reinforcement learning, in the context of edge learning.}
    \item \textbf{Explore tools and libraries}: {We survey tools and libraries for training ML models on edge devices, as well as simulations and emulators for edge learning.}
    \item \textbf{Use-cases and applications}: {We present various use cases and applications of edge learning explored in academic research.}
\end{enumerate}

Table~\ref{tab:edgeml_related_surveys} present the relevant studies related to Edge ML, and compare them to our survey based on the aforementioned {topics. The symbols used in the table convey the extent to which each study addresses the different topics of edge learning outlined previously.}

\begin{enumerate}
    \item {~{\cmark} indicates that a survey comprehensively covers a particular topic in the context of edge learning.}
    \item {$\circ$ denotes partial coverage of the topic, where a study may focus on a specific subset of the topic. For example, the surveys~\cite{tak_federated_2021, abreha_federated_2022, boobalan_fusion_2022} explores techniques for using/optimizing ML for the edge but only focus on federated learning.}
    \item {~{\textbullet} signifies that a survey touches on the topic, but its primary emphasis lies on inference on the edge, rather than training, which often results in less comprehensive coverage of the training aspects.}
    \item {~{\xmark} indicates that a study does not address the topic at all.}
\end{enumerate}

\begin{table*}[t!]
\caption{Summary of Edge Machine Learning related surveys}
\resizebox{\textwidth}{!}{%
\newcolumntype{C}{>{\centering\arraybackslash}X}

\begin{tabularx}{\fulllength}{p{3.5cm}CCCCCCC}

\toprule
\textbf{Survey} & \textbf{Year} & \textbf{Explore techniques} & \textbf{Metrics for Edge Learning} & \textbf{Compare techniques} & \textbf{Explore Types of ML} & \textbf{Explore tools and libraries} & \textbf{Use-cases and applications}\\
\midrule

Chen et al.~\cite{Chen2019DeepLearningEdge}
& 2019 & \textbullet & \xmark & \textbullet & \xmark & \textbullet & \textbullet \\
\midrule

Wang et al.~\cite{wang_convergence_2020}
& 2020 & \textbullet & \xmark & \textbullet & \xmark & \textbullet & \textbullet \\
\midrule

Shi et al.~\cite{shi_communication-efficient_2020}
& 2020 & \textbullet & \xmark & \textbullet & \xmark & \xmark & \xmark \\
\midrule

{Xu et al.}~\cite{xu2020edge}
& 2020 & \textbullet & \xmark & \textbullet & \xmark & \xmark & \xmark \\
\midrule

{Lim et al.}~\cite{lim_federated_2020}
& 2020 & $\circ$ & \xmark & \xmark & \xmark & \xmark & \xmark \\
\midrule

Leon Veas et al.~\cite{10.1007/978-3-030-68080-0_6}
& 2021 & \textbullet & \xmark & \xmark & \xmark & \xmark & \xmark \\
\midrule

Dhar et al.~\cite{Dhar2021SurveyOnDevice}
& 2021 & $\circ$ & \xmark & $\circ$ & \xmark & \xmark & \xmark \\
\midrule

Tak et al.~\cite{tak_federated_2021}
& 2021 & $\circ$ & \xmark & \xmark & \xmark & \xmark & \xmark \\
\midrule

{Zhang et al.}~\cite{zhang_edge_2021}
& 2021 & \textbullet & \cmark & \xmark & \xmark & \cmark & \xmark \\
\midrule

Murshed et al.~\cite{murshed_machine_2022} 
& 2022 & \textbullet & \xmark & \xmark & \xmark & \textbullet & \textbullet \\
\midrule

Abreha et al.~\cite{abreha_federated_2022}
& 2022 & $\circ$ & \xmark & \xmark & \xmark & $\circ$ & $\circ$ \\
\midrule

Boobalan et al.~\cite{boobalan_fusion_2022} 
& 2022 & $\circ$ & \xmark & \xmark & \xmark & \xmark & $\circ$ \\
\midrule

Joshi et al.~\cite{joshi_enabling_2022}
& 2022 & \cmark & \cmark & \cmark & \xmark & \xmark & \xmark \\
\midrule

Cai et al.~\cite{cai_enable_2022}
& 2022 & \textbullet & \xmark & \textbullet & \xmark & \xmark & \xmark \\
\midrule

Cui et al.~\cite{cui_federated_2022}
& 2022 & $\circ$ & \xmark & \xmark & \xmark & \xmark & $\circ$ \\
\midrule

Imteaj et al.~\cite{imteaj_survey_2022}
& 2022 & $\circ$ & \xmark & $\circ$ & \xmark & \xmark & $\circ$ \\
\midrule

{Mendez et al.}~\cite{mendez_edge_2022}
& 2022 & \xmark & \xmark & \xmark & \xmark & \textbullet & \xmark \\
\midrule

{Filho et al.}~\cite{filho_systematic_2022}
& 2022 & \textbullet & \xmark & \xmark & \xmark & \textbullet & \textbullet \\
\midrule

{Ray et al.}~\cite{ray_review_2022}
& 2022 & \xmark & \xmark & \xmark & \xmark & \textbullet & \xmark \\
\midrule

Li et al.~\cite{li_review_nodate} 
& 2023 & \textbullet & \textbullet & \cmark & \xmark & \textbullet & \xmark \\
\midrule

Hua et al.~\cite{hua_edge_2023}
& 2023 & \textbullet & \xmark & \xmark & \xmark & \xmark & \textbullet \\
\midrule

Zhu et al.~\cite{zhu_-device_2023}
& 2023 & $\circ$ & \cmark & $\circ$ & \xmark & \xmark & \xmark \\
\midrule

Wu et al.~\cite{wu_survey_2023}
& 2023 & $\circ$ & \xmark & $\circ$ & \xmark & \xmark & \xmark \\
\midrule

{Hoffpauir et al.}~\cite{hoffpauir_survey_2023}
& 2023 & \xmark & \xmark & \xmark & \xmark & \xmark & \textbullet \\
\midrule

{Barbuto et al.}~\cite{Barbuto_disclosing_2023}
& 2023 & \textbullet & \xmark & \xmark & \xmark & \xmark & \xmark \\
\midrule

{Trinade et al.}~\cite{trinade_Resource_2024}
& 2024 & $\circ$ & \xmark & \xmark & \xmark & \xmark & \xmark \\
\midrule

{Grzesik et al.}~\cite{grzesik_combining_2024}
& 2024 & \xmark & \textbullet & \xmark & \xmark & \textbullet & \textbullet \\
\midrule

{Jouini et al.}~\cite{Jouini_Machine_2024}
& 2024 & \xmark & \xmark & \xmark & \xmark & \textbullet & \textbullet \\
\midrule

Our survey
& 2024 & \cmark & \cmark & \cmark & \cmark & \cmark & \cmark \\
\bottomrule

\end{tabularx}
}
\noindent{\footnotesize{\cmark : {Fully covers the topic of Edge Learning};\\ $\circ$ : {Partially covers the topic of Edge Learning};\\ \textbullet : {Focuses on both Edge Learning and Inference}; \\ \xmark : {Does not cover edge learning at all};}}

\label{tab:edgeml_related_surveys}

\end{table*}

\subsection{{Structure of the survey}}

This survey is {organized into six main sections}, excluding this introduction and the conclusion (Section~\ref{sec:conclusion}). First, {Section~\ref{sec:edge-comp-learning} provides a detailed definition of edge computing, edge learning and edge devices, and present the requirements and metrics for training ML models at the edge}. In Section~\ref{sec:overview-techs}, we explore the techniques used to enable, optimize, and accelerate edge learning. {A detailed comparison between these techniques is presented in Section~\ref{sec:comparison-techniques}}. {In Section~\ref{sec:types}, we discuss the integration of different types of ML such as unsupervised learning or reinforcement learning in the edge, to leverage these techniques in edge learning, optimize their performance, or enable the training of other models}. In Section~\ref{sec:use-case}, we explore the use cases and current applications of edge learning. Then in Section~\ref{sec:libs-and-tools}, we present different tools, frameworks and libraries used to {create simulations and} train ML models at the edge. Finally, {Section~\ref{sec:challenges} identifies open challenges and discusses potential future trends and research directions in edge learning}.

\section{Edge computing and Edge learning}
\label{sec:edge-comp-learning}
{This section introduces the fundamental concepts of edge computing, edge machine learning, and edge learning. We will then examine the essential requirements of edge learning.}

\subsection{Edge Computing}
\label{sec:edge-comp}
Edge computing is a new computing paradigm that aims to address the limitations of traditional cloud computing models in handling large scale data generated by the increasing number of smart devices connected to the Internet. It involves performing calculations at the edge of the network, closer to the user and the source of the data. Edge computing emphasizes local, small-scale data storage and processing, providing benefits such as reduced bandwidth load, faster response speed, improved security, and enhanced privacy compared to traditional cloud computing models~\cite{Cao2020EdgeComputingOverview}.

Edge computing addresses several limitations of cloud computing, that stem from the frequent communications needed between end/edge devices and cloud server, in the standard cloud computing paradigm and the reliance of storing data centrally, which might compromise the privacy or security of sensible data.

\begin{itemize}
    \item \textbf{Reduced latency}: Edge computing brings data processing closer to the source, reducing the time it takes for data to travel to a centralized cloud server, thereby reducing latency and improving response time~\cite{ayyasamy2023edge}.
    \item \textbf{Bandwidth {reduction}}: Edge computing reduces the need for transmitting large amounts of data to centralized cloud servers, resulting in reduced bandwidth load and reduced network congestion~\cite{Cao2020EdgeComputingOverview}.
    \item \textbf{Improved data privacy}: Edge computing allows for local data processing, reducing the need to transmit sensitive data to centralized cloud servers, thereby minimizing the risk of data breaches and unauthorized access~\cite{Cao2020EdgeComputingOverview}.
    \item \textbf{Operational resilience}: Edge computing enables applications to continue functioning even in disconnected or low-bandwidth environments, ensuring operational resilience and reducing dependency on centralized cloud infrastructures~\cite{ayyasamy2023edge}.
\end{itemize}

\begin{figure}
    \centering
    \includegraphics[scale=0.5]{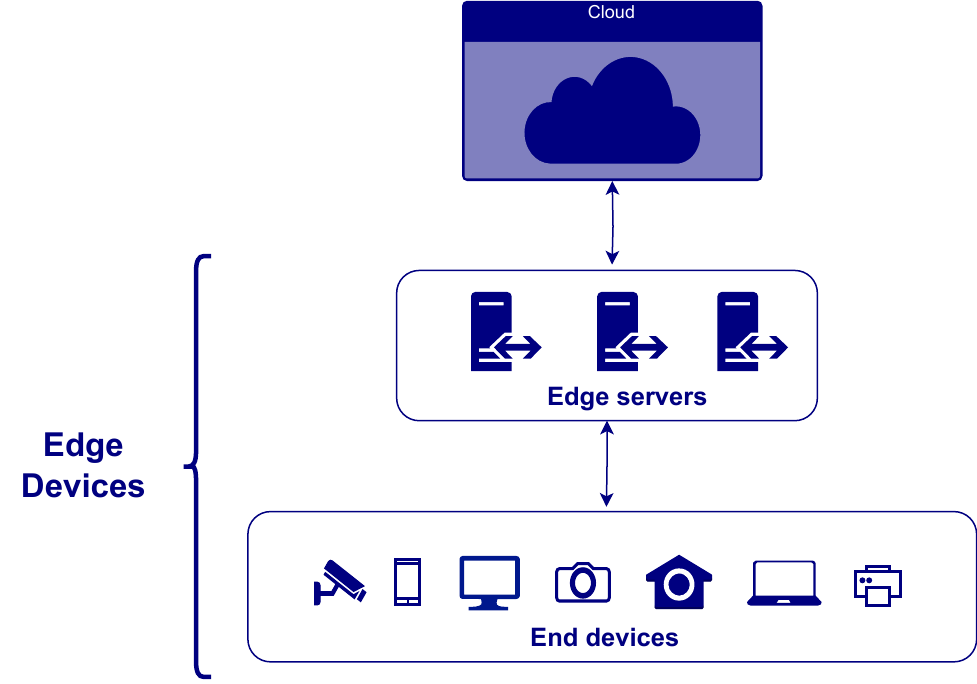}
    \caption{A typical architecture of edge computing}
    \label{fig:edge-computing-archi}
\end{figure}

Figure~\ref{fig:edge-computing-archi} shows the general architecture of edge computing, inspired by the ones proposed in~\cite{Chen2019DeepLearningEdge} and~\cite{varghese2016ChallengesEdge}. We define edge devices as both edge servers and end devices, as well as other types of devices that weren't specifically mentioned in the diagram, such as routers and routing switches. {For a deeper dive into edge computing, the reader can consult to edge computing surveys such as~\cite{liu_survey_2019, Cao2020EdgeComputingOverview}}.

\subsection{Edge Learning}
\label{sec:edge-learning}
{A key advancement in edge computing is the integration of AI and ML. Edge ML enables the training and deployment of ML models directly on edge devices, which includes both edge learning and edge inference. Edge learning, also known as edge training, involves training ML models directly on edge devices, reducing the reliance on centralized cloud infrastructure. In contrast, edge inference focuses on facilitating the inference of ML models on resource-constrained edge devices, regardless of where the models were trained}~\cite{lu2020scaling}. {Another term commonly used in the literature is edge intelligence, which shares similarities with edge ML. However, it also includes data collection, caching, processing, and analysis at the edge, making it a broader concept than edge ML}~\cite{xu2020edge}.

{While} most Edge ML research focuses on edge inference~\cite{baller_deepedgebench_2021, wu2019machine}, {edge learning remains a promising approach. By enabling localized model training}, edge learning can be tailored to the specific requirements and resource constraints of edge devices, making it ideal for applications that require privacy preservation and model customization for specific use cases.

Edge Learning employs various strategies, most of them are either categorized as distributed or collaborative learning methods, {which} distribute the training of ML models across multiple edge devices, such as federated or split learning; and on-device learning, {which involves training} ML models on individual edge devices, {and may employ optimization or fine-tuning techniques as necessary}.

In this survey, we will explore both on-device learning and distributed learning on edge devices. Distributed learning is defined as the training of ML models collaboratively across multiple devices. {In contrast,} on-device learning refers to the training of ML models in a single device. To ensure clarity, our definition of edge devices also encompasses edge servers, network elements, and end devices. We comprehensively address ML model training across all these devices.

\subsection{{Requirements for edge learning}}\label{sec:requirements}
The successful training of ML models at the edge requires meeting specific requirements that dictate the {efficiency} and performance of these models. These requirements are essential for ensuring that the models perform optimally and efficiently within the resource-constrained environment inherent to edge devices. While there is no single metric to define the efficiency of training ML models in the edge~\cite{bailey_edge_2022}, {different ones could be constructed be used to evaluate if and how well the aforementioned requirements are met, and estimate the performances of the model in resource's contained environment.}

\begin{enumerate}
    \item \textbf{Computational Efficiency}: Computational Efficiency refers to the ability of an algorithm to achieve high performance with minimal computational cost. This is especially important in the context of {edge learning}, as edge devices {often} have limited computational resources~\cite{baller_deepedgebench_2021}, and ML models typically require high computational complexity {for their training}~\cite{kearns_computational_1990}.
    \item \textbf{Memory Footprint Efficiency}: Similarly to computational complexity, edge devices often have low memory availability~\cite{cai_tinytl_2020, lin_-device_2022}, {which} contrast with the large memory {requirements} of ML models.
    \item \textbf{Fast Training Time}: Fast training time {refers} to the rapid convergence of model parameters during the training phase. {Fast training time is crucial for edge devices, as it directly impacts their efficiency and responsiveness}. {Edge devices are often characterized by limited computational capabilities, as mentioned earlier}. As such, they require ML models to be trained swiftly to minimize the processing burden and reduce energy consumption. {Fast training time also enables models to adapt quickly to changing data patterns, ensuring responsiveness and adaptivity. This allows models to be efficiently updated to address dynamic environments and changing user requirements}.
    \item {\textbf{Minimized Bandwidth Consumption}}: {Reducing bandwidth usage involves minimizing data transfer between edge devices and improving communication efficiency among them. This} is particularly important for distributed learning techniques and especially in bandwidth-limited systems, since these techniques require frequent sharing of the ML model across the network devices.
    \item \textbf{Low Energy Consumption}: Energy consumption is a crucial consideration {for edge devices}, especially in mobile edge computing. {This is due to the} limited energy available on such devices. {Therefore}, ML models trained at the edge {must be} energy-efficient {to ensure better computing performance, longer battery life, and successful model training}. Energy efficiency refers to the ability {to perform tasks or functions using minimal energy}. It involves reducing energy waste and optimizing energy consumption.
    \item \textbf{Labelled Data Independency}: {Most edge-generated data is unlabelled}~\cite{10012038}. {Therefore}, {using ML techniques that can handle unlabeled data}, such as unsupervised (Section~\ref{sec:unsupervised-learning}), self-supervised (Section~\ref{sec:self-supervised-learning}), or semi-supervised learning (Section~\ref{sec:semi-supervised-learning}), {may be beneficial in edge learning.}
    \item \textbf{Task Specific Metrics and Performance}: {Since} edge learning encompass different ML tasks and use-cases. Specific metrics and benchmarks are {commonly used to evaluate a model's performance to assess} its effectiveness in achieving its intended goals.
\end{enumerate}

\section{Overview of Edge Learning techniques}\label{sec:overview-techs}

{In general, edge ML training is similar to traditional ML training, with the added requirements and constraints outlined in Section~\ref{sec:requirements}. The feasibility} of training on the edge depends on the resource requirements of the model, and the {device resources}. {Today, the increasing processing power, energy storage, and memory capacity of edge devices}~\cite{Afachao2022ARO} {enables small ML models to be trained on edge devices without requiring significant optimization or distribution}. For example, KMeans in~\cite{bellotti_exploring_2021}, Self-Organizing Map in~\cite{zhu_evaluation_2021} and SVM in~\cite{Yazici2018EdgeMachineLearning}. However, the training of more complex models that require heavier resources, such as neural networks, is more challenging at the edge. Therefore, in this section, we will present an overview of techniques used to optimize the training of more complex ML models on the edge. {These techniques include distributing training across multiple devices, cloud-based training with local fine-tuning, and model optimization or compression to enable edge training}.

Figure~\ref{fig:edge-learning-techs-overview} shows a global view of edge learning techniques reviewed in this paper. The techniques are separated into four categories: (i) Distributed or collaborative techniques, such as federated or split learning; (ii) Techniques that rely on fine-tuning of a model trained on the cloud, such as Transfer or incremental learning; (iii) techniques that compress models to facilitate or support the training on the edge, such as quantization and knowledge distillation; (iv) And finally the other optimization techniques that don't fit {neatly} into the previous categories.

\definecolor{custom_blue}{rgb}{0, 0.7490196078431373, 0.8196078431372549}
\definecolor{CustomDarkBlue}{rgb}{0, 0, 0.5019607843137255}

\tikzset{
    my node/.style={
        draw=CustomDarkBlue,
        outer color=CustomDarkBlue,
        color=CustomDarkBlue,
        thick,
        minimum width=1cm,
        text width=35ex,
        text height=0ex,
        text depth=0ex,
        font=\sffamily,
        drop shadow,
    }
}

\begin{figure*}[t]
    \centering
    \begin{forest}
        forked edges,
        for tree={%
            my node,
            rounded corners,
            l sep+=10pt,
            grow'=east,
            align=center,
            minimum height=0.65cm,
            edge={gray, thick},
            parent anchor=east,
            child anchor=west,
            top color=white,
            bottom color=white,
            edge path={
                \noexpand\path [draw, \forestoption{edge}] (!u.parent anchor) -- +(10pt,0) |- (.child anchor)\forestoption{edge label} [CustomDarkBlue];
            },
            if={isodd(n_children())}{
                for children={
                    if={equal(n,(n_children("!u")+1)/2)}{calign with current}{}
                }
            }{}
        },
        [Edge Learning, text width=17ex
            [Distributed and\\ collaborative learning (\ref{sec:distributed-training-on-edge}), text depth=2ex, text width=27ex, minimum height=1cm
                [Federated Learning (\ref{sec:federated-learning})]
                [Split Learning (\ref{sec:split-learning})]
                [Swarm Learning (\ref{sec:swarm-learning})]
                [Gossip Learning (\ref{sec:gossip-learning})]
            ]
            [Adaptive and fine-tuning \\ based technique (\ref{sec:Fine-tuning-training-on-edge}), text depth=2ex, text width=27ex, minimum height=1cm)
                [Transfer Learning (\ref{sec:transfer-learning})]
                [Incremental Learning (\ref{sec:incremental-learning})]
                [Meta Learning (\ref{sec:meta-learning})]
            ]
            [Model compression (\ref{sec:compression-training-on-edge}), text width=27ex, minimum height=1cm
                [Quantization (\ref{sec:quantization})]
                [Knowledge Distillation (\ref{sec:knowledge-distillation})]
                [Model Pruning (\ref{sec:pruning})]
            ]
            [Other optimization \\ methods (\ref{sec:optimization-training-on-edge}), text depth=2ex, text width=27ex, minimum height=1cm
                [Binary Neural Networks (\ref{sec:bnn})]
                [Spiking Neural Networks (\ref{sec:snn})]
                [Forward-Forward Algorithm (\ref{sec:forward-forward})]
            ]
        ]
    \end{forest}
    \caption{{A taxonomy of techniques used to enable and/or optimize edge learning}}\label{fig:edge-learning-techs-overview}
\end{figure*}
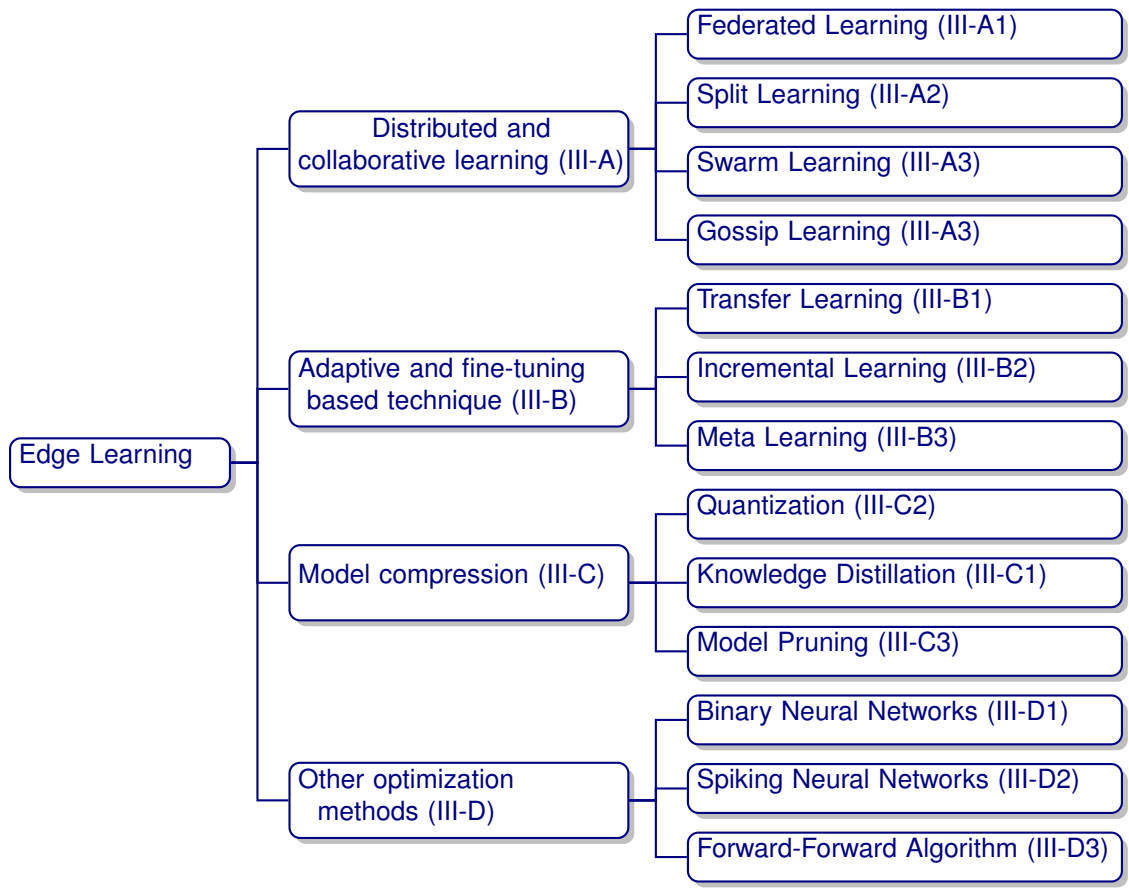

\subsection{Distributed and Collaborative Techniques}
\label{sec:distributed-training-on-edge}
In this section, we will explore distributed techniques to train {ML} models at the edge. They work by leveraging the computational capabilities of multiple edge devices, and aggregating their results, instead of relying on a single resource constrained device.

\subsubsection{Federated Learning}
\label{sec:federated-learning}

Federated Learning (FL), offers a transformative approach to decentralized model training. In the context of edge learning, where data is distributed across numerous edge devices, FL enables collaborative training without centralizing sensitive data~\cite{mcmahan_communication-efficient_2016}. This technique involves training a shared model across these devices by iteratively updating it based on local data, {with the objective of preserving data privacy~\cite{das_privacy_2019, jiang_privacy-preserving_2023, liu_privacy-preserving_2022}. FL has been widely adopted for edge learning}~\cite{abreha_federated_2022}, with applications in various domains, including cyberattack detection~\cite{abbas_novel_2023,li_predicting_2021}, spam detection~\cite{sidhpura_fedspam_2023,sriraman_-device_2022}, smart cities~\cite{el_hanjri_federated_2023, shi_towards_2023}, and autonomous vehicles~\cite{cheriguene_towards_2024}.

In order to train an FL algorithm, an aggregation method is needed. Federated learning aims to generate a global model by aggregating local models from multiple clients. This process combines individual models to create a generalized one that represents the collective knowledge of all clients. The main two aggregation methods being, {Federated Stochastic Gradient Descent (FedSGD) and Federated Averaging (FedAVG)}~\cite{mcmahan_communication-efficient_2016}. However, other approaches have been proposed over the years such as EdgeFed~\cite{ye_edgefed_2020} {which reduces FedAvg's computational overhead} by separating the process of updating the local model that is supposed to be completed independently by mobile devices, or FedSel~\cite{liu_fedsel_2020} {which addresses FedSGD's dimension dependency} problem, by selecting Top-k dimensions according to their contributions in each iteration. Other approaches include MTFeeL~\cite{mahara_multi-task_2022}, FedDynamic~\cite{zeng_adaptive_2023}, FedNets~\cite{alhalabi_fednets_2023}, FedCom~\cite{zhao_fedcom_2023}, FedGPO~\cite{kim_fedgpo_2022}, and FedOVA~\cite{liu_resource-constrained_2022}.

Despite its growing popularity and multiple benefits, traditional FL models suffer from some limitations. For instance, Non-IID (Non-Independent and Identically Distributed) data often negatively impacts the performance of the global model~\cite{zeng_adaptive_2023}. FL is also {vulnerable} to malicious and low-quality users~\cite{lv_data-free_2021}, emerging new classes with completely unseen data distributions whose data cannot be accessed by the global server or other users~\cite{gudur_zero-shot_2021} as well as single node failure~\cite{Savazzi2020FLCooperating, Gupta2022FedFM}, channel bandwidth bottlenecks~\cite{agrawal_decentralizing_2020}, and scaling issues for increasing network size~\cite{Savazzi2020FLCooperating}. To solve the low performance with Non-IID data challenge, Hybrid {FL} approaches have been proposed~\cite{gu_fedaux_2022}, where very small amounts of data is shared between participants. Other approaches that aim to solve this problem include FedNets~\cite{alhalabi_fednets_2023}, FedDynamic~\cite{zeng_adaptive_2023} and~\cite{zhang_optimizing_2021} which proposes a one-shot neural architecture search technique. {In contrast,} pairwise correlated agreement~\cite{lv_data-free_2021}, is a method that aims to evaluate individual users' contribution to avoid malicious and low-quality contributions from users. Sharma et al.~\cite{sharma_federated_2022} proposes a framework to study different noise patterns in user feedback, and explore noise-robust mitigation techniques for training FL models. {Finally,}~\cite{gudur_zero-shot_2021} proposes a unified zero-shot framework to handle emerging classes in edge devices.

Hierarchical federated learning{, an extension of FL, introduces} a multi-level architecture~\cite{liu_client-edge-cloud_2020}, {which enables} more efficient communication and computation trade-offs~\cite{abad_hierarchical_2020}. {Furthermore, they facilitate} faster model training and reduce energy consumption by offloading tasks to edge servers for partial model aggregation to reduce network traffic~\cite{ma_data-aware_2022}. For instance,~\cite{wen_towards_2022} proposes a hierarchical training algorithms that address challenges in helper scheduling and communication resource allocation. {While} ~\cite{ma_data-aware_2022} {developed} a task offloading approach based on data and resource heterogeneity to improve training performance and reduce system cost. Other variations of FL include blind federated edge learning~\cite{amiri_blind_2021}, modular federated learning~\cite{liang_modular_2022}, and clustered federated learning~\cite{sattler_clustered_2019} which will be presented in more details in section~\ref{sec:unsupervised-learning}.

{There is a growing interest in training language and multimedia models at the edge using FL, with several approaches being proposed in recent years}. While the training of Large Language Models (LLMs) using FL is still experimental, some approaches have been proposed such as FATE-LLM~\cite{fan2023fate} and FwdLLM~\cite{xu_federated_2023} which aim to fine-tune {a billion parameter} language models across mobile devices using FL. {In contrast, relatively} smaller language models like BERT~\cite{devlin2018bert} {have been extensively explored using FL, with approaches} such as FedBERT~\cite{tian_fedbert_2022} that uses FL and SL approaches for pre-training BERT in a federated way. FedSplitBERT~\cite{li_federated_2022} {addresses the challenges of} heterogeneous data and decreases the communication cost by splitting BERT encoder layers into {two parts. A local part trained on the client-side} and a global part trained by aggregating gradients of multiple clients. {Another example is} FedSPAM~\cite{sidhpura_fedspam_2023}, {which fine-tunes} a distilBERT model~\cite{sanh2019distilbert} using FL on mobile devices to detect spams in SMSes. In computer vision, FedVKD~\cite{tao_training_2022} was proposed as a federated knowledge distillation training algorithm to train small CNN models on edge devices. {Periodically, the knowledge from these models is transferred to a} large server-side vision transformer encoder via knowledge distillation. {On the other hand,}~\cite{hsu_federated_2020} introduces an FL approach for visual classification with real-world data distribution. Finally, training audio models at the edge using FL is wildly explored for tasks such as speech recognition~\cite{yu_federated_2021, jia_federated_2022, guliani_enabling_2022, guliani_training_2021, bai_robust_2021} or audio classification~\cite{gudur_zero-shot_2021, zhang2023fedaudio}.

{One important concept usually related to FL is differential privacy. Differential privacy is a privacy preservation technique that involves adding artificial noise to protect individual privacy while maintaining model utility~\cite{hidayat_agc-dp_2023}.~\cite{dwork_differential_2008} provides a detailed examination of differential privacy while~\cite{zhang_systematic_2023} and~\cite{weiFederatedLearningDifferential2020} provides an exploration of differential privacy in the context of FL.}

{FL} is also used alongside other techniques presented in this survey, such as split learning~\cite{thapa_splitfed_2022}, meta learning~\cite{zou_fedmc_2022, yue_efficient_2022}, transfer learning~\cite{suzuki_using_2021}, knowledge distillation~\cite{wu_survey_2023, tanghatari_federated_2023, qu_quantization_2020}, and Quantization~\cite{liu_hierarchical_2023, qu_quantization_2020, li_federated_2022}, etc.

\subsubsection{Split Learning}
\label{sec:split-learning}
Split learning {offers an alternative approach to} collaborative learning. In contrast to FL, which involves training models on local data from different devices and aggregating them on a central server, {split learning takes a different approach. Specifically,} it divides the model into sections, with each section trained on a different client {or server, and} instead of transferring raw data, only the weights of the last layer of each section are sent to the next client. This process ensures model improvement while maintaining better data and model privacy than FL, {thanks to} the model architecture split between clients and the server. {Additionally,} this split makes split learning a more suitable option for resource-constrained environments, where computational resources are limited. {However, this approach comes at the cost of} slower processing than FL, which is due to its relay-based training~\cite{thapa_splitfed_2022}.

{In recent years, there has been a growing interest in split learning at the edge, as evident from the increasing number of studies in this area} (see figure~\ref{fig:papers_per_technique_and_year}). For instance, SplitEasy~\cite{palanisamy_spliteasy_2021} is a framework that enables the training ML models on mobile devices using {split learning}. Another paper,~\cite{liu_masking-enabled_2023} proposes a data protection approach for {split learning} without compromising the model accuracy. Additionally,~\cite{fu_joint_2023_split} proposes an online model splitting method with resource provisioning game scheme which aims to minimize the total time cost of participating devices. Adaptive {split learning}, is a branch of {split learning} that aims to overcome its shortcomings compared to FL. {Specifically, it} addresses these challenges by eliminating the transmission of gradients from the server to the client, resulting in a smaller payload and reduced communication cost, and allowing the client to update only sparse partitions of the server model, adapting to the variable resource budgets of different clients which decreases the computation cost and improves performance across heterogeneous clients~\cite{chopra_adaptive_2023, chopra_adasplit_2021}. Other adaptative approaches include ARES~\cite{samikwa_ares_2022} and~\cite{ayad_improving_2021}. Finally, {split learning} has been combined with FL {in multiple approaches} in order to eliminate both techniques' inherent drawbacks, notably in~\cite{thapa_splitfed_2022, cheng_cheese_2023, duan_combined_2022, li_high-precision_2023, fu_privacy-preserving_2023, tao_training_2022, zhang_cluster-hsfl_2023}.

\subsubsection{Other Collaborative Learning methods}\label{sec:swarm-learning}\label{sec:gossip-learning}
Several distributed learning techniques have been proposed as alternatives to FL and {split learning}. Swarm learning, an innovative approach integrating artificial and biological intelligence, addresses challenges in distributed ML for the edge. This method efficiently utilizes signal processing and communication techniques to operate in real-time within large-scale edge IoT environments, offering advantages in overcoming communication bottlenecks, diverse data, non-convex optimization, and privacy concerns~\cite{wang_distributed_2022}. Another approach to swarm learning, CB-DSL~\cite{fan_efficient_2023}, a Communication-efficient and Byzantine-robust Distributed Swarm Learning technique, was introduced to deal with Non-IID data issues and byzantine attacks. Another noteworthy distributed learning method is gossip learning, which, like other collaborative methods, doesn't require transferring data outside edge devices. However, unlike FL and other methods, gossip learning operates without a central server for model aggregation and lacks reliance on central control~\cite{hegedus_gossip_2019}. Notable extensions to gossip learning, such as the one proposed by~\cite{bagoly_round_2020}, enhance the algorithm by incorporating additional memory for storing local caches of model updates, making it more suitable for mobile devices.
\subsection{Adaptive and Fine-tuning based Techniques}\label{sec:Fine-tuning-training-on-edge}
In this section, we discuss techniques for efficiently adapting and fine-tuning pre-trained ML models at the edge without requiring a complete retraining. The focus of these techniques is on preserving privacy and achieving personalized performance, while reducing computational overhead by keeping the heaviest part of the training in the cloud or edge servers, either using public datasets or ethically collected data. We explore three approaches: transfer learning, incremental learning, and meta learning. These methods enable edge devices to leverage previously acquired knowledge, adapt to local data distributions, and the continuous improvement of the models on the edge.

\subsubsection{Transfer Learning}\label{sec:transfer-learning}
Transfer learning is an ML technique where knowledge gained from solving one problem is applied to a different, yet related, problem. Instead of building models from scratch, {transfer learning} employs pre-trained models on large datasets to extract valuable insights, such as learned features or representations. These insights are then used to enhance the performance of a new task, especially when limited data is available for that task. By capitalizing on existing knowledge, {transfer learning} accelerates model training, improves generalization, and proves exceptionally useful in domains where data scarcity poses a challenge~\cite{pan2009survey}. {For a more detailed understanding of transfer  learning, readers are encouraged to review the surveys~\cite{tan_survey_2018, zhuang_comprehensive_2021, pan2009survey}.}

In the context of edge learning, {transfer learning} is a prominent technique used to fine-tune ML models based on local data in an edge device. This approach serves as a fully on-device alternative to collaborative learning methods that distribute the training across different devices~\cite{lin_-device_2022}. {Notable state-of-the-art methods for transfer learning in edge learning include tiny-transfer learning}~\cite{cai_tinytl_2020}, {which addresses the critical issue of memory efficiency in low-memory edge devices. This is achieved} this by freezing the weights of the model and only learning a memory-efficient bias module, thus removing the need to store the intermediate activations. Similarly, RepNet~\cite{yang_repnet_2022} proposes an intermediate feature re-programming of a pre-trained model with a tiny reprogramming network to develop memory-efficient on-device transfer learning. MobileTL~\cite{chiang_mobiletl_2023} also proposes a memory and computationally efficient on-device transfer learning method, {specifically designed for models built with inverted residual blocks}. Additionally,~\cite{yang_joint_2022} propose an edge CNN framework for 5G industrial edge networks, with the CNN model trained in advance in an edge server, which is further fine-tuned based on the limited datasets uploaded from the devices with the aid of transfer learning, and~\cite{choi2022accelerating} proposes a runtime convergence monitor to achieve massive computational savings in the practical on-device training workloads. Multiple approaches also focus on combining {transfer learning with} FL, to create federated transfer learning algorithms~\cite{ahmed_federated_2021}, that aim to leverage FL for privacy preservation, and use transfer learning to train a well-performing local model despite users usually having not enough data for that by training the base model with a public dataset and passing it to the federated users to be fine-tuned for the target task~\cite{suzuki_using_2021, liu_fedct_2021}. Finally,~\cite{vucetic_efficient_2022} proposes freeze and reconfigure, a {transfer learning} method for on-device training of a BERT model.

\subsubsection{Incremental Learning}\label{sec:incremental-learning}
Incremental learning also known as continual learning or life-long learning, involves continuously updating and expanding a model's knowledge as new data becomes available. Unlike traditional batch learning, where models are trained from scratch on entire datasets, incremental learning dynamically incorporates new information without discarding previously acquired knowledge~\cite{wu_large_2019}, and can be used to reduce/overcome the well-known issue of catastrophic forgetting in deep neural networks~\cite{zuo_handling_2023, shi_overcoming_2021, doan_efficient_2023}. {Readers seeking a more thorough understanding of incremental learning are directed to~\cite{yang2019survey}.}

There have been considerable attempts of implementing incremental learning in the context of edge learning. These include: learning with sharing~\cite{hussain_learning_2023} which aims to reduce the training complexity and memory requirements while achieving high accuracy during the incremental learning process and bypass the considerable memory requirements that can make incremental learning unsuited for edge devices; PILOTE~\cite{zuo_handling_2023} which trains an incremental learning model on edge devices for human activity recognition;~\cite{disabato_incremental_2020} introduces an incremental algorithm based on transfer learning and k-nearest neighbor to support the on-device learning; RIANN~\cite{liu_riann_2020} is an indexing and search system for graph-based approximate nearest neighbor algorithm for mobile devices; and RILOD~\cite{li_rilod_2019} which aims to incrementally train an existing object detection model to detect new object classes without losing its capability to detect old classes, to avoid catastrophic forgetting. RILOD distills three types of knowledge from the old model to mimic the old model's behaviour on object classification, bounding box regression and feature extraction, and it was implemented under both edge-cloud and edge-only setups~\cite{li_rilod_2019}. There are a variety of promising approaches and directions for incremental learning on the edge from combining it with other techniques (such as {FL}~\cite{rao_federated_2023, yue_inexact-admm_2021}, meta learning~\cite{luo_learning_2020, yue_inexact-admm_2021} and compression methods~\cite{wang_efficient_2022, carta_projected_2023, zhang_impact_2021}) to sparse~\cite{wang_sparcl_2022} or distributed continual learning~\cite{carta_projected_2023}.

\subsubsection{Meta-learning}\label{sec:meta-learning}
Meta learning focus on enhancing a model's ability to learn new tasks quickly and effectively. Unlike traditional learning paradigms that optimize for a specific task, meta learning trains models to learn from a diverse set of tasks, thereby enabling them to generalize knowledge and adapt rapidly to novel tasks with minimal data~\cite{nichol_first-order_2018}. By exposing models to various learning scenarios, meta learning equips them with transferable skills, such as recognizing patterns and adapting to new contexts. {For further insight into meta learning, we recommend consulting~\cite{huisman_survey_2021, hospedales_meta-learning_2020}.}

In the context of edge learning, the application of meta learning introduces a transformative approach to address the challenges posed by limited data availability and resource constraints~\cite{qu_p-meta_2022, rosenfeld_fast_2021}.~\cite{gao_pruning_2021} proposes adaptation-aware network pruning, a model pruning method designed to work with existing meta learning methods to achieve fast adaptation on edge devices, while~\cite{luo_learning_2020} proposes a continual meta-learning approach with bayesian graph neural networks that mathematically formulates meta-learning as continual learning of a sequence of tasks, and p-Meta was introduced in~\cite{qu_p-meta_2022}, and aims to achieve faster generalization to unseen tasks and enforces structure-wise partial parameter updates to support memory-efficient adaptation. Meta learning can also be used in combination with other techniques such as {FL}~\cite{yue_inexact-admm_2021, yu_communication-efficient_2023}, or~\cite{zou_fedmc_2022} which integrates reinforcement learning models trained by multiple edge devices into a general model based on a meta-learning approach, in order to create FedMC, a generalized federated reinforcement learning framework based on a meta-learning approach.

\subsection{Model Compression based Techniques}
\label{sec:compression-training-on-edge}
This section explores model compression techniques, which aim to streamline the training of ML models at the edge.  As traditional deployment and inference solutions have embraced knowledge distillation, quantization, and model pruning to accelerate model execution on resource-constrained devices, a notable shift is observed towards employing these techniques for reducing the complexity of ML models for the training in the edge, making these techniques helpful for the training phase as well.

\subsubsection{Knowledge distillation}
\label{sec:knowledge-distillation}
Knowledge distillation in deep learning is a process whereby a small or student neural network is trained to emulate the knowledge and predictive capabilities of a larger or teacher network. This technique serves as a means to transfer the expertise and generalization capabilities of a complex model to a simpler one{. As a result, inference efficiency is enhanced, and computational demands are reduced}. The underlying principle involves the student network learning not only from ground truth labels but also from the soft, probabilistic outputs of the teacher network, thereby capturing finer details and nuances in the data~\cite{gou2021knowledge}. {For a deeper dive into knowledge distillation, readers can refer to the following survey ~\cite{gou2021knowledge}}. In the context of edge learning, knowledge distillation is usually used to reduce the size and complexity of a large neural network, to simplify its training in limited resources devices. Therefore, knowledge distillation is well suited to be used in collaboration with other techniques such as federated learning~\cite{wu_survey_2023, tanghatari_federated_2023}, split learning~\cite{nam_active_2023} or incremental learning~\cite{wang_efficient_2022}. However, distillation is also used independently of other techniques~\cite{xia_-device_2022, qian_two-stage_2022}.

{The integration of knowledge distillation with FL on edge devices has shown promising results, with recent trends indicating great potential in combining the two techniques}~\cite{wu_survey_2023}. {Several approaches have been proposed, including attack-resistant FL methods}~\cite{zhou_communication-efficient_2023}, speech recognition tasks~\cite{bai_robust_2021} or keyword spotting~\cite{hard_production_2022}. Mix2FLD~\cite{oh_mix2fld_2020} is another method that {combines knowledge distillation} and FL. {Meanwhile,}~\cite{qu_quantization_2020} use both {distillation} and quantization to train {FL} models on edge devices. {Several other hybrid approaches combining FL and knowledge distillation have been explored}~\cite{ahn_wireless_2019, nguyen_enhancing_2023}{. These approaches are further discussed in Wu et al.'s survey on knowledge distillation in federated edge learning}~\cite{wu_survey_2023}. {Knowledge distillation can also be applied to other distributed learning methods, such as}~\cite{nam_active_2023}, which introduces a spatio-temporal distillation method for split learning for a tiny server in order to alleviate the frequent communication costs that happen when communicating from the server to edge devices.~\cite{bistritz_distributed_2020} introduces a distributed distillation algorithm where devices communicate and learn from soft-decision outputs, which are inherently architecture-agnostic and scale only with the number of classes in order to alleviate the communication costs from transmitting model weights in the network and improve the inclusion of devices with different model architectures. Finally, {knowledge distillation} has been used as a standalone technique in an edge learning context, for recommendation systems~\cite{xia_efficient_2023, yao_device-cloud_2021, xia_-device_2022}, edge cardiac disease detection~\cite{Wong2023CardiacDisease} and on-device deep reinforcement learning~\cite{jang_knowledge_2020}. {Knowledge distillation} was, additionally, used with multiple variants, including dataset distillation techniques~\cite{sebti_dataset_2024, accettola_dataset_2023} and knowledge transfer~\cite{hu_variational_2022, qian_two-stage_2022}.

\subsubsection{Quantization}
\label{sec:quantization}
Quantization in deep learning refers to the process of reducing the precision of numerical values representing model parameters or activations, typically from floating-point to fixed-point or integer representations, in order to balance the act of maintaining an acceptable level of model accuracy while significantly reducing the memory and computational requirements~\cite{gholami2022survey}. This computational optimization technique is pivotal in mitigating the resource-intensive demands of deep neural networks, rendering them more amenable for resource-constrained hardware platforms, such as edge devices and embedded systems. There are two types of quantization: quantization-aware training and post-training quantization. In {quantization-aware training}, the quantized model is fine-tuned using training data in order to adjust parameters and recover accuracy degradation or perturbation introduced by the quantization. {In contrast, post-training quantization is a less expensive approach, where the pretrained model is quantized and its weight adjusted without any fine tuning}~\cite{gholami2022survey}. {To gain a more complete understanding of quantization techniques, readers are advised to consult~\cite{gholami2022survey}}

Quantization techniques are not only used to optimize machine learning models before deployment for edge inference~\cite{kwasniewska_deep_2019, choi2022accelerating}, but also in edge learning to simplify the fine-tuning of large models~\cite{lin_-device_2022, ostertag_efficient_2020}. Similarly to {knowledge distillation}, quantization is often used alongside other techniques such as FL~\cite{noauthor_optimization_2023}, transfer Learning~\cite{lin_-device_2022}, incremental learning~\cite{choi_optimized_2019} or with other types of techniques~\cite{chen_3u-edgeai_2021}. Quantization is used with FL particularly extensively, including one-bit quantization~\cite{li_one_2023, zhu_one-bit_2021, li2022federated} and hierarchical FL~\cite{liu_hierarchical_2023}{. Other FL-based approaches that utilize quantization include}~\cite{cui_communication-efficient_2023, noauthor_joint_2022, ren_research_2023, chen_energy_2023-1, liu_training_2022-1, qu_quantization_2020}. Other quantization-based methods for training ML models in the edge include quantization-aware scaling, which was proposed in~\cite{lin_-device_2022}{. This method automatically scales} the gradient of tensors with different bit-precisions without requiring any fine-tuning, and was used alongside a tiny training engine and sparse updates. {Investigations in}~\cite{chauhan_exploring_2022} {showed that quantization helps further in reducing the resource requirements for the training on on-device few shot learning for audio classification. The Holmes optimizer}~\cite{yamagishi_holmes_2022} {uses quantization to improve the accuracy by combining different quantization techniques, such as limiting quantization bits, fixed-point numbers, and logarithmic quantization.}

\subsubsection{Model Pruning}
\label{sec:pruning}
Model pruning is a technique used to reduce the size of ML models by removing certain parts of the model, such as model parameters, nodes in a decision tree~\cite{zhou2019model} or weight matrices in transformer-based models~\cite{kwon2022fast}. Similarly to quantization, model pruning is commonly used in the edge to reduce the computational resources required for the inference of ML models~\cite{choi2021convergence}. {Model pruning has also shown great potential in edge learning, where it reduces the size of ML model before fine-tuning on edge devices. This technique is particularly effective when used in conjunction with other methods, such as FL}~\cite{ren_research_2023, yu_heterogeneous_2023, jiang_model_2023, liu_joint_2022}, incremental learning~\cite{disabato_incremental_2020, zhang_impact_2021} or meta-learning~\cite{gao_pruning_2021}.

Similar to knowledge distillation and quantization, {FL} emerges as the most prominent technique when combined with model pruning for edge learning. Noteworthy is PruneFL~\cite{jiang_model_2023}, an approach aimed at minimizing communication and computation overhead while reducing training time through adaptive model size adjustment during FL. PruneFL employs model pruning, starting with an initial pruning stage at a selected client, followed by subsequent pruning iterations during FL. {Other approaches that combine model pruning with FL include}~\cite{liu_joint_2022} {which introduces} model pruning for wireless {FL} to scale down neural networks{. Meanwhile,}~\cite{yu_heterogeneous_2023} employ an adaptive dynamic pruning approach to prevent overfitting by slimming the model through the dropout of unimportant parameters. {In addition, several approaches use model pruning on edge devices. For instance,}~\cite{mairittha_-device_2021} uses model pruning in the context of on-device personalization for an activity recognition system, and Deeprec~\cite{han_deeprec_2021} leverages model pruning and embedding sparsity techniques to reduce computation and network overhead. Furthermore, OmniDRL~\cite{lee_omnidrl_2022}, a deep reinforcement learning based approach on edge devices, incorporates weight pruning in each learning iteration to achieve a high weight compression ratio. Finally,~\cite{hosny_sparse_2021} explores the reduction in memory footprint for further pruning during the training phase of BitTrain, a bitmap memory efficient compression technique for training on edge devices.
\subsection{Optimization and Acceleration based Techniques}\label{sec:optimization-training-on-edge}
In this section, we explore some other techniques that {don't fit neatly} into a specific category and are used to optimize or provide more optimized alternatives to machine learning models' {enabling them to be} more suitable for edge learning.

\subsubsection{Binary neural networks}\label{sec:bnn}
Binary Neural Networks (BNNs) are deep neural networks that use binary values (-1 or 1) instead of floating-point numbers for weights and activations. BNNs are attractive for resource-constrained devices because of their ability to compress deep neural networks~\cite{courbariaux2016binarized}. {BNNs share similarities with other techniques, such as quantization and model pruning, which are also considered good} candidates for edge inference due to their extreme compute and memory savings over higher-precision alternatives~\cite{wang_enabling_2023}. However, {BNNs' compute and memory efficiency can also be leveraged for edge learning. For example, by proposing a hybrid quantization of a continual learning model}~\cite{vorabbi_-device_2023}, {or by developing a model based on an MRAM array with ternary gradients for both training and inference on the edge}~\cite{fujiwara_bnn_2023}. Other BNN based approaches for edge learning include~\cite{penkovsky_-memory_2020, pham_efficient_2021, wang_enabling_2023}.

\subsubsection{Spiking neural networks}\label{sec:snn}
Spiking neural networks (SNNs) are another type of deep neural networks that {are} promising for the edge. SNNs communicate between neurons using events called spikes~\cite{gerstner2002spiking} and are known for their asynchronous and sparse computations. {These properties result in decreased energy consumption}~\cite{tang2014energy, lemaire2022analytical}, {which makes them well suited for energy limited devices}~\cite{xue2023edgemap}. Training ML models at the edge using SNNs has gained some {attention in recent years. For example,}~\cite{skatchkovsky_federated_2020} proposes FL-SNN, a cooperative training through FL for networked on-device SNNs, while~\cite{zyarah_-device_2018} presents a memristor spiking neuron and synaptic trace circuits for efficient on device learning. {Other approaches include integrating meta-learning with SNNs for lifelong learning on a stream of tasks with local backpropagation-free nested updates}~\cite{rosenfeld_fast_2021}{, and using event-driven, power and memory-efficient local learning rules, such as spike-timing-dependent plasticity}~\cite{soures_-device_2017}. There are other approaches that leverage SNNs for edge learning, including~\cite{stratton_unsupervised_2023, tang_seneca_2023, safa_neuromorphic_2022}.

\subsubsection{Forward-Forward Algorithm}\label{sec:forward-forward}
The backpropagation algorithm is essential for training neural networks, but recent studies have proposed alternatives to the algorithm when the available resources are limited. {One such algorithm is the} forward-forward~\cite{hinton_forward-forward_2022} algorithm, which replaces the forward and backward passes of backpropagation by two forward passes that operate in the same way as each other on different data with opposite objectives. A positive pass operates on real data and adjusts the weights to increase the goodness in every hidden layer, while a negative pass operates on "negative data" and adjusts the weights to decrease the goodness in every hidden layer~\cite{hinton_forward-forward_2022}. {To adapt the forward-forward algorithm to edge devices, researchers have proposed variations such as µ-FF}~\cite{de_vita_-ff_2023}, {which uses} a multivariate ridge regression approach and allows finding closed-form solution by using the mean squared error. {Another study}~\cite{pau2023suitability} investigates the improvements in terms of complexity and memory usage brought by PEPITA~\cite{pmlr-v162-dellaferrera22a} and the forward-forward algorithm. the results show that the forward-forward algorithm reduces memory consumption by 40\% on average, but involves additional computation at inference that, can be costly on microcontrollers.

\subsubsection{Other techniques}
In this section, we will explore some other techniques used to optimize ML models for training in the edge. One such technique is data booleanization, used in~\cite{rahman_data_2022}, which proposes a novel approach towards low-energy booleanization. MiniLearn~\cite{profentzas_minilearn_2022} on the other hand, enables re-training of deep neural networks on resource-constrained IoT devices{. This allows them} to re-train and optimize pre-trained, quantized neural networks using IoT data collected during deployment. {Another approach is the use of echo state networks for anomaly detection in aerospace applications}~\cite{carta_efficient_2023}. Tiny training engine~\cite{lin_-device_2022} is a lightweight training system, introduced alongside sparse update, a technique that skip the gradient computation of less important layers and sub-tensors, and a quantization-aware scaling to stabilize 8-bit quantized training. Tiny training engine enables on-device training of convolutional neural networks under 256KB of SRAM and 1MB flash without auxiliary memory~\cite{lin_-device_2022}.

\cite{nadalini_reduced_2023} introduces a novel reduced precision optimization technique for on-device learning primitives on MCU class devices with a specialized shape transform operators and matrix multiplication kernels, {which is} accelerated with parallelization and loop unrolling for the backpropagation algorithm. {In addition,} POET~\cite{patil_poet_2022} allows for the training of large neural networks on memory scarce and battery-operated edge devices, with integrated rematerialization and paging. POET reduce the memory consumption of backpropagation, {allowing the fine-tuning of} both ResNet-18 and BERT within edge devices' memory constraints. Finally,~\cite{chen_3u-edgeai_2021} proposes a novel rank-adaptive tensor-based tensorized neural network mode for on-device {training} with ultra-low memory usage.
\subsection{Comparison of techniques used in edge learning}\label{sec:comparison-techniques}
In this section we will compare the different families of techniques used to train ML models in the edge that we explored in the previous part of this paper (sections~\ref{sec:distributed-training-on-edge},~\ref{sec:Fine-tuning-training-on-edge},~\ref{sec:compression-training-on-edge},~\ref{sec:optimization-training-on-edge}), we will compare the different families based on two factors: (i) The number of academic contributions of each family and their evolution over the years; (ii) The techniques' potential of answering the different needs and requirements particular {to} edge learning.

\subsubsection{Comparison of the usage of the different techniques}
\begin{figure*}[t]
    \centering
    \includegraphics[scale=0.7]{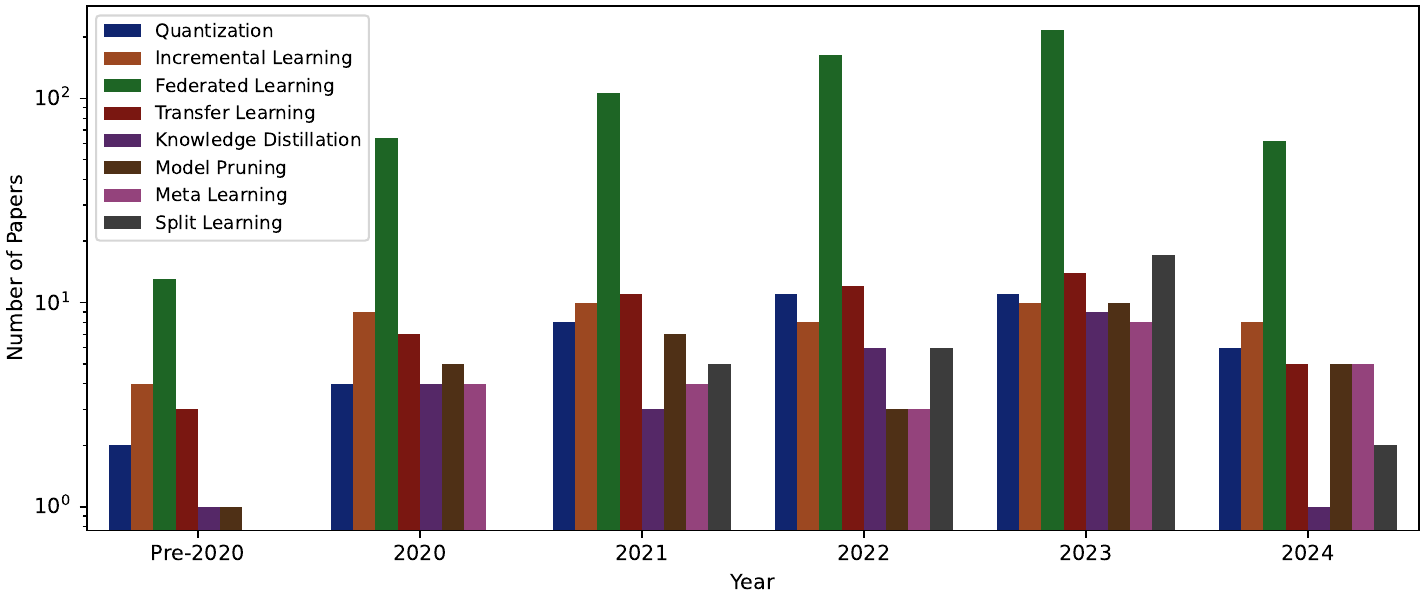}
    \caption{Trend of techniques used to train ML models in the edge over the years}\label{fig:papers_per_technique_and_year}
\end{figure*}

\begin{figure}[t]
    \centering
    \includegraphics[scale=0.55]{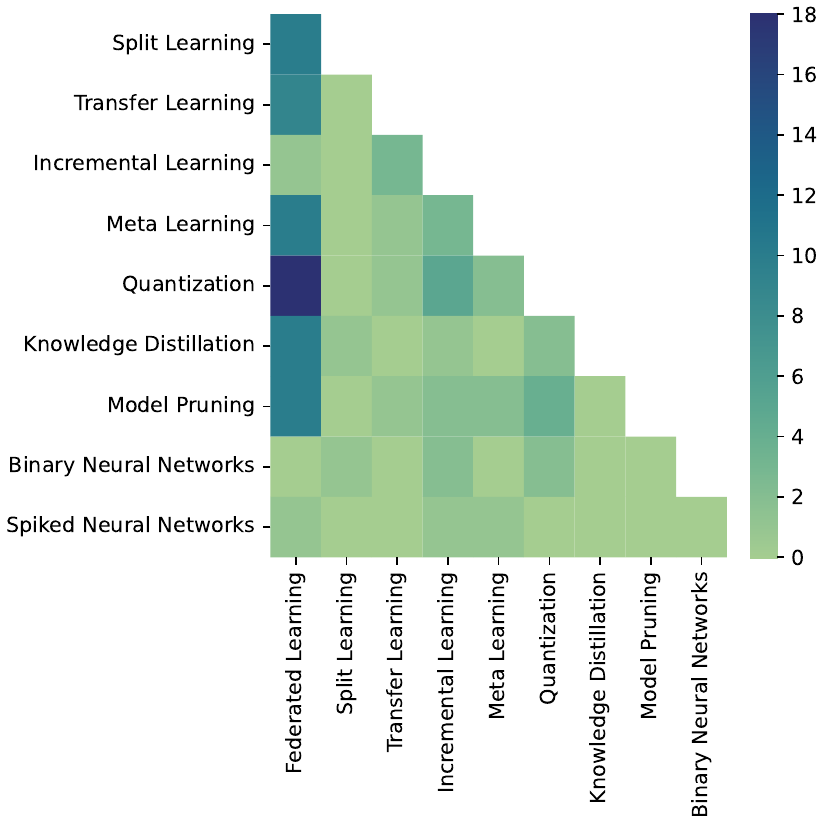}
    \caption{Trend of different techniques used hand in hand for training ML models at the edge. Color intensity represents the number of papers}
    \label{fig:heatmap-techniques}
\end{figure}

We will first start {by comparing the different edge learning techniques} over the years by analyzing their academic contributions. Figure~\ref{fig:papers_per_technique_and_year} {shows the number of papers per technique per year on a logarithmic scale}. {We used the advanced search features} of Scopus\footnote{Scopus: \url{https://www.scopus.com/}} {and Web of Science\footnote{Web of Science: \url{https://clarivate.com/products/scientific-and-academic-research/research-discovery-and-workflow-solutions/webofscience-platform/}} to get the data}. {Using those search engines, we searched for the terms} "edge learning", {"Edge Intelligence",} "training/learning on the edge/mobile devices", "on-device training/learning" and "on-device adaptation" as well as relevant keywords for each technique ("Federated Learning", "Split Learning", etc.), {in the title, keywords and abstract. We excluded} surveys, books, and notes, as we are only interested {in technical contributions to optimizing ML model training on the edge} using the aforementioned techniques. {Additionally, we manually reviewed each paper} and removed papers that either didn't provide a technical contribution, or weren't about training ML models {on the edge}, despite containing relevant keywords. Finally, {we added a few manually found papers that were not indexed in both Scopus and Web of Science or did not contain the relevant keywords, but were still relevant to our analysis}. {Note that we excluded the families of techniques that had less than 10 papers in total for edge learning. The excluded techniques {are:} swarm learning, gossip learning, forward-forward, BNNs and SNNs.} The final number of papers associated with the analysis was {803} papers, and some of these papers were briefly covered in~\ref{sec:overview-techs}. Note that multiple techniques can be used in a single paper {(see Figure~\ref{fig:heatmap-techniques})}, therefore the total count of techniques {in the Figure}~\ref{fig:papers_per_technique_and_year} will exceed {803. The cut-off date for the year 2024 is the 20th July 2024.}

The analysis of Figure~\ref{fig:papers_per_technique_and_year} reveals that {FL} is the dominant approach for training ML models in edge environments, given the resource constraints of edge devices. This dominance is expected, as distributed learning methods that capitalize on the collective computing power of multiple devices are deemed more practical and efficient in edge settings. Moreover, we anticipate that this trend will persist and expand to include {split learning}, another promising distributed learning technique. Other methods, including incremental learning, transfer learning, Model Compression Techniques (e.g., quantization, knowledge distillation), although consistently employed, lag behind FL in terms of popularity. A closer examination of Figure~\ref{fig:papers_per_technique_and_year} unveils a remarkable surge in the number of publications focused on edge learning over the past six years. Notably, there has been a steady rise in the adoption of FL and split learning, aligning with our forecast of a trend favoring these two techniques. In contrast, the use of techniques like incremental learning and transfer learning has been more steady during the same period, and meta-learning despite being employed consistently {in previous years} received less attention in 2023. Finally, for the model compression techniques while quantization and {knowledge distillation} experienced a small rise in popularity over the years, model pruning has exhibited greater fluctuation during that period.

As previously discussed, various approaches have been proposed that integrate multiple techniques to mitigate the limitations of individual methods and capitalize on their respective strengths. Examples of such approaches include~\cite{thapa_splitfed_2022, yue_inexact-admm_2021, luo_learning_2020, ahmed_federated_2021}. To provide a clearer illustration of the relationships between these various techniques, a heatmap depicting the intersection of their usage is presented in Figure~\ref{fig:heatmap-techniques}. This visual representation allows for a more comprehensive understanding of the synergies and overlap between different approaches. {The heatmap includes all the technique families discussed in Section~\ref{sec:overview-techs}, except for swarm learning, gossip learning, and the forward-forward algorithm, which have not been combined with other techniques in the context of edge learning to our knowledge.} We can note that FL has been combined the most with other techniques, which is expected considering the overwhelming number of FL contributions to the edge (see Figure~\ref{fig:papers_per_technique_and_year}). {Furthermore}, {model compression} techniques such as {knowledge distillation}, model pruning and quantization are also {often} used together with other techniques as discussed in Sections (\ref{sec:knowledge-distillation}~\ref{sec:quantization}~\ref{sec:pruning}). Finally, we can {observe} other collaborations between the different families, and we expect this trend to continue in the future.

\subsubsection{Comparison based on the requirements and needs of edge learning} \label{sec:comparison-based-requirements}
As {outlined} in Section~\ref{sec:requirements}, there are several requirements that must be met for edge learning, which can function as imprecise measures for assessing the viability of the families of techniques covered in earlier sections {for the training in the edge}. {Upon reevaluating the requirements, we excluded} {“labelled data independence”} from the comparison, as it is more related to the type of {ML employed or} the availability of an autolabeling process {rather than to the techniques being evaluated}. {Furthermore, we do not consider “task-specific metrics and performance” as this encompasses multiple metrics that vary depending on the specific task at hand}. {However, we introduce a “high performance”} measure to estimate roughly if the strategies positively or negatively affect the performances. For instance, model compression techniques often lower model performance, whereas incremental and meta-learning typically boost it. {As a result, we use six distinct measures to evaluate the techniques:} {computational efficiency, memory footprint, low energy consumption, quick training time, reduced bandwidth, and high performance}. Table~\ref{tab:techniques_comparison} compares the various families of techniques against these requirements.

The outcomes, as depicted in Table~\ref{tab:techniques_comparison}, should be interpreted as informal and approximate assessments aimed at providing a broad understanding of the general strengths and weaknesses of the compared techniques within the context of edge learning. A checkmark ("\cmark") implies that, in general, the technique offers assistance or advantages when applied to {edge learning} with respect to the specified requirement. Conversely, a cross mark ("\xmark") indicates that, in general, the requirement represents a weakness of the technique in the {edge learning} context. It is important to note that techniques like FL, quantization, and others have various variants and specific approaches that influence how these methods align with the requirements. For instance, while quantization techniques may result in a minor decrease in performance in most scenarios~\cite{Choukroun2019LowBitQuantization}, leading to an "\xmark" check in the "High Performance" column. Certain specific approaches to quantization may exceptionally yield no performance loss or even improvements, as demonstrated in studies such as~\cite{yamagishi_holmes_2022}. Therefore, it is important to note that the assessment provided by these symbols should be considered as rough estimates, as the effectiveness of a technique, and how it fares against a specific requirement, can vary depending on diverse factors such as variant versions, implementation details, use cases, tasks, and hardware platforms. Accordingly, table~\ref{tab:techniques_comparison} offers a general overview rather than a definitive judgment on the suitability of each technique for every situation.

\begin{table*}[t!]
\caption{Comparison between The different techniques that enable edge learning}
\label{tab:techniques_comparison}
\resizebox{\textwidth}{!}{%

\newcolumntype{C}{>{\centering\arraybackslash}X}
\begin{tabularx}{\fulllength}{p{4.5cm}CCCCCC}

\toprule
\textbf{Technique} & \textbf{Computation Efficiency} & \textbf{Memory footprint} & \textbf{Low energy consumption} & \textbf{Fast Training time} & \textbf{Optimized Bandwidth} & \textbf{High Performance} \\

\midrule
Federated Learning
& \cmark & \xmark & \cmark & \cmark & \textbullet & \xmark \\

\midrule
Split Learning
& \cmark & \cmark & \cmark & \xmark & \xmark & \textbf{\textbackslash} \\

\midrule
Swarm Learning
& \cmark & \textbf{\textbackslash} & \textbf{\textbackslash} & \textbf{\textbackslash} & \cmark & \textbf{\textbackslash} \\

\midrule
Gossip Learning
& \cmark & \xmark & \xmark & \textbf{\textbackslash} & \xmark & \textbf{\textbackslash} \\

\midrule
Transfer Learning
& \textbullet & \xmark & \textbullet & \textbullet & \cmark & \cmark \\

\midrule
Incremental Learning
& \textbullet & \xmark & \textbullet & \textbullet & \cmark & \cmark \\

\midrule
Meta-Learning
& \textbullet & \xmark & \textbullet & \textbullet & \cmark & \cmark \\

\midrule
Knowledge Distillation
& \cmark & \cmark & \cmark & \cmark & \cmark & \xmark \\

\midrule
Quantization
& \cmark & \cmark & \textbullet & \cmark & \cmark & \xmark \\

\midrule
Model Pruning
& \cmark & \cmark & \cmark & \cmark & \cmark & \xmark \\

\midrule
BNNs
& \cmark & \cmark & \cmark & \cmark & \cmark & \xmark \\

\midrule
SNNs
& \cmark & \textbf{\textbackslash} & \cmark & \textbf{\textbackslash} & \textbf{\textbackslash} & \xmark \\

\midrule
Forward-Forward Algorithm
& \cmark & \cmark & \cmark & \cmark & \textbf{\textbackslash} & \textbf{\textbackslash} \\

\bottomrule
        
\end{tabularx}
}

\noindent{\footnotesize{\\
\cmark: Have a positive effect on the requirement;\\
\xmark: Have a negative effect on the requirement;\\
\textbullet: Have a neutral or uncertain effect based on specific conditions on the requirement;\\ 
\textbf{\textbackslash}: There is not enough information and literature to estimate the effect on the requirement}}

\end{table*}

{Overall, analyzing the results from Table~\ref{tab:techniques_comparison}, we can observe some high level trends, distributed techniques like FL and split learning exhibit computational efficiency due to their inherent distributed nature, reducing the load on individual devices. In contrast, we observe differences in memory footprint, with FL performing poorly due to its requirement for loading the entire model on each device, whereas split learning only requires loading a portion of the model. However, split learning is less efficient in terms of bandwidth usage and training time, as it necessitates frequent transmission of output from different splits. Similarly, gossip learning's decentralized nature makes it less optimal for bandwidth usage, whereas swarm learning offers advantages in overcoming communication bottlenecks, reducing bandwidth usage. In terms of performance, FL often yields lower results than centralized alternatives, although this is not universally true across all approaches. It is worth noting that even when two techniques meet a requirement, they may not do so with the same level of efficiency. For instance, both FL and split learning meet the computational efficiency requirement, but split learning may offer greater efficiency, particularly when dealing with large models. Transfer learning, incremental learning, and meta-learning have all characteristics in common. Although they do not optimize for memory, as the model typically needs to be fully loaded for training, they often result in improved performance and reduced bandwidth usage compared to distributed methods. However, their impact on other requirements is generally less clear. Some approaches significantly optimize for these measures, while others don't. Finally, model compression techniques inherently optimize for memory, computation, and energy requirements by reducing model complexity. As a result, they also reduce bandwidth usage compared to non-compressed models, simply by decreasing the model size. However, these methods often result in decreased performance.}

{To conclude with the analysis of Table~\ref{tab:techniques_comparison} and Figures~\ref{fig:papers_per_technique_and_year} and~\ref{fig:heatmap-techniques}. Despite some drawbacks, FL has established itself as a cornerstone technique for edge learning, with successful adaptations across various domains and tasks. Moreover, combining FL with other techniques or specific implementations can mitigate its performance limitations and reduce memory and bandwidth usage. On the other hand, split learning shows great potential when combined with FL and is particularly promising for larger models, and we anticipate further advancements in this area. In contrast, adaptive and fine-tuning-based techniques are often a great choice when cloud pretraining is possible, reducing the amount of training needed on the edge, enabling further model personalization, and showing great potential when combined with distributed techniques. Model compression techniques are well-suited for edge devices, as they reduce model size, thereby decreasing computational, memory, and energy consumption. However, this often comes at the cost of decreased performance. Ultimately, each technique has its strengths, and the choice of technique should be based on the specific task and constraints at hand. Furthermore, combining multiple techniques can be beneficial, as it allows leveraging their individual strengths.}

\section{Edge Learning for different types of Machine Learning}\label{sec:types}
In this section, we will explore the usage of different types of ML in the edge. We will focus on unsupervised learning, reinforcement learning, semi-supervised and self-supervised learning, as they present some particularities when adapted to the edge, however, we will ignore supervised learning~\cite{mohsin_fpga-based_2018} as it's usually considered the default when it comes to model training and most approaches at the edge use it without requiring any adaptation or particular implementation.

\subsection{Unsupervised Learning}\label{sec:unsupervised-learning}
Considering the vast amount of unlabeled data produced in edge and end devices~\cite{10012038}, it is very promising to use unlabeled data to train ML models on the edge. However, unsupervised learning comes with multiple challenges and restrictions for edge learning, especially when it comes to collaborative learning techniques such as FL or {split learning}, which represent the vast majority of techniques used in the edge (see Figure~\ref{fig:papers_per_technique_and_year}). Unsupervised learning datasets may have a {non-IID} nature. Each node in a collaborative setting might have a different subset of the data, and the data distribution might vary across nodes. This non-IID property can make it difficult to effectively combine information from different nodes. Additionally, in the case of clustering algorithms, clusters may have varying sizes across nodes in a collaborative setting and clustering algorithms may need to adapt to changes in data distribution and cluster structures over time. Finally, since there are no available labels, and their assignment may differ between nodes (for example in a clustering algorithm). Ensuring consistency across distributed nodes is difficult, but crucial for aggregating meaningful global labels (clusters).

\begin{figure}[]
  \centering
  \includegraphics[scale=0.55]{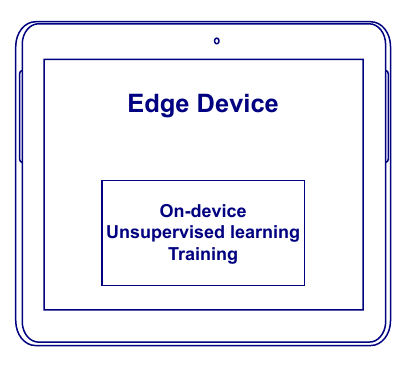}
  \caption{{Unsupervised learning with the training happening only on a single device, with all the required data hosted locally}}
  \label{fig:ondevice_unsupervised}
\end{figure}

\begin{figure}[]
  \centering
  \includegraphics[scale=0.55]{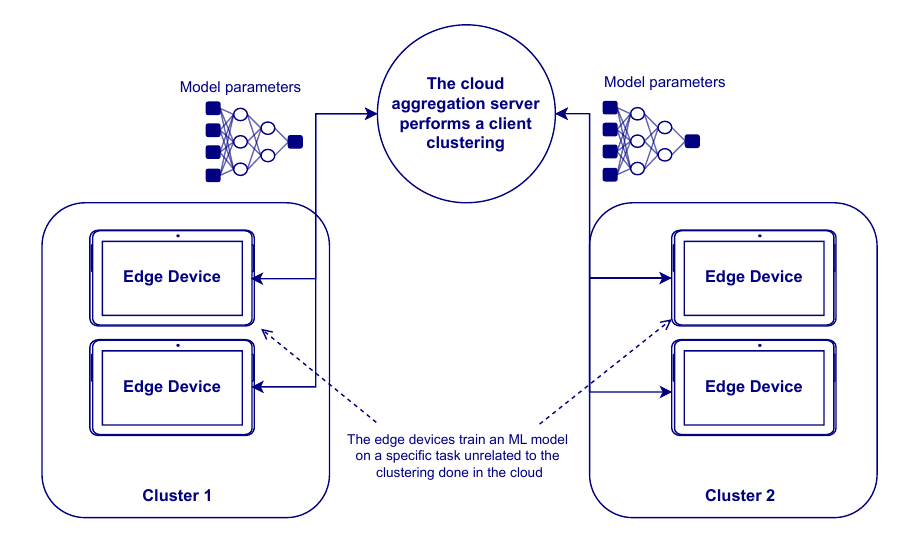}
  \caption{{Unsupervised learning to assist collaborative learning, a clustering is typically applied to edge devices to improve on the FL process (For example CFL)}}
  \label{fig:assist_unsupervised}
\end{figure}

\begin{figure}[]
  \centering
  \includegraphics[scale=0.55]{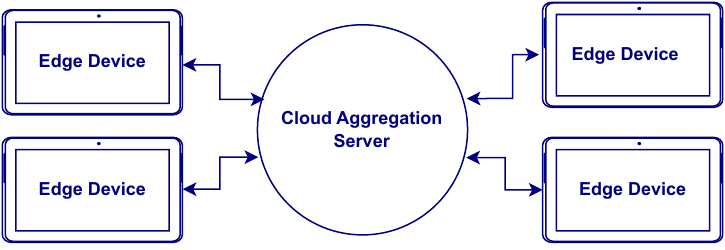}
  \caption{{Collaborative unsupervised learning, the goal being to apply unsupervised learning algorithms on completely distributed data}}
  \label{fig:collaborative_unsupervised}
\end{figure}

Figures~\ref{fig:ondevice_unsupervised},~\ref{fig:assist_unsupervised} and~\ref{fig:collaborative_unsupervised} represent the main different types of unsupervised learning approaches used for training ML models in the edge, (a) using unsupervised learning algorithms directly on-device with non-collaborative methods~\cite{yang_-device_2023, zhu_evaluation_2021} as shown in Figure~\ref{fig:ondevice_unsupervised}; (b) using unsupervised learning methods to assist in the training of collaborative learning approaches, such as clustered federated learning~\cite{sattler_clustered_2019, albaseer_client_2021, wang_clustered_2022} highlighted in Figure~\ref{fig:assist_unsupervised}; and (c) training an unsupervised learning model on the edge collaboratively such as federated clustering~\cite{dennis_heterogeneity_2021} in Figure~\ref{fig:collaborative_unsupervised}.

\paragraph{Unsupervised learning on a single edge device} Having an unsupervised learning model trained on a single edge device is possible if the device has enough computation power and/or the learning algorithm is lightweight and can be trained with low resources, the training with such models is usually no different from the training on the cloud or other devices, the only difference being the constraint of low resources and data available~\cite{zhu_evaluation_2021}. Examples of unsupervised learning algorithms trained with this approach include~\cite{bellotti_exploring_2021} which investigates the application of K-Means on mainstream controllers, and~\cite{hsieh_cyclegan_2020} that presents the first dedicated CycleGAN accelerator for energy-constrained mobile applications, achieving a higher throughput-to-area ratio and higher energy efficiency than a GPU. In~\cite{yang_-device_2023}, an unsupervised segmentation was proposed that can be executed on edge devices without the need of annotated data. While~\cite{muthu_unsupervised_2022}, proposes TMNet, an approach to solve  unsupervised video object segmentation problem at the edge. Finally,~\cite{piyasena_dynamically_2020} proposes an FPGA based architecture for a self-organization neural network capable of performing unsupervised learning on input features from a CNN by dynamically growing neurons and connections in order to perform class-incremental lifelong learning for object classification in the edge.

\paragraph{Unsupervised learning to assist collaborative learning approaches} Collaborative learning approaches, such as FL, are promising solutions for training ML models on edge devices. {However, FL, the most popular technique on edge devices, faces multiple challenges, including non-IID data, uneven computing power}~\cite{li_high-precision_2023} and suboptimal results when the local clients' data distributions diverge~\cite{sattler_clustered_2019}. {To address these issues, the usage of clustering alongside FL have been proposed multiple times to cluster devices with similar environmental data distributions}~\cite{li_hpfl-cn_2022}. {One popular approach is} Clustered Federated Learning (CFL)~\cite{sattler_clustered_2019}, {which exploits geometric properties of the FL loss surface and group the client into clusters with jointly trainable data distributions.} {CFL has been widely adopted and has inspired other approaches, such as}~\cite{albaseer_client_2021, li_high-precision_2023, wang_clustered_2022, gong_towards_2023}. {Alternative approaches include hierarchical over-the-air FL}~\cite{aygun_hierarchical_2022}, {which utilizes intermediary servers to form clusters near mobile users}, and HPFL-CN~\cite{li_hpfl-cn_2022}, a communication-efficient hierarchical personalized {FL framework that uses complex network feature clustering to group edge servers with similar environmental data distributions}. Subsequently, personalized models are trained for each cluster using a hierarchical architecture, resulting in enhanced efficiency. HPFL-CN incorporates privacy-preserving feature clustering to derive low-dimensional feature representations for each edge server. This is achieved by mapping the environmental data onto various complex network domains, thereby accurately clustering edge servers with similar characteristics. Finally,~\cite{gudur_zero-shot_2021} uses unsupervised learning methods on the edge to distinguish between classes across different users, when new classes with completely unseen data distributions emerge on devices in {an FL} setting for audio classification.

\paragraph{Collaborative Unsupervised learning at the edge} {Collaborative unsupervised learning methods are challenging to train for all the reason explained previously.} Nevertheless, several methodologies have surfaced. {Among them is federated clustering, which aim to execute clustering on distributed data without sharing the data}~\cite{dennis_heterogeneity_2021, yan_ccfc_2024, yan_ccfc_2024-1}. {Federated clustering methods can be described as one-shot if they require only one round of transfer between clients and the servers}~\cite{dennis_heterogeneity_2021, yan_privacy-preserving_2023}, {or they can require multiple round of communication}~\cite{stallmann_towards_2022}. An alternative methodology, known as FedUReID, has been proposed by Zhuang et al.~\cite{zhuang_joint_2021} as a person Re-identification system without the use of any labels, all the while ensuring the preservation of privacy. {Finally, federation of unsupervised learning}~\cite{lu_federated_2022} {proposes a method where unlabeled data undergo a transformation process to become surrogate labelled data for each client. Following this, a modified model is trained through supervised FL. Eventually, the desired model is obtained by recovering it from the modified model.}

\subsection{Reinforcement learning}\label{sec:reinforcement-learning}
Reinforcement learning (RL) has been successfully applied in the past on different problems in areas such as robotics, recommendation systems, video games and automatic vehicles~\cite{RLStudy2023Sivamayil, IntroRL2020Naeem}, making RL a promising and interesting direction for edge learning. However, The training of RL models in resources constrained environments is often limited by high compute and memory requirements from gradient computations~\cite{kao_e3_2021}, making the application of RL in the context of edge learning challenging. Despite these challenges, multiple RL approaches have been proposed for training ML models in the edge. Among them, {federated RL}~\cite{zhuo2020federated, Qi_2021} is a promising approach that allow multiple RL agents to learn optimal control policies for a series of devices with slightly different dynamics~\cite{xianjia_federated_2021}. {Furthermore, it} is employed to achieve diverse objectives including personalization~\cite{Nadiger2019123, Xiong2024}, IoT traffic management~\cite{jarwan_edge-based_2023, FDRL2020Wang}, Autonomous Systems~\cite{xianjia_federated_2021} and resource allocation for unmanned aerial vehicle (UAV)~\cite{liu_deep_2023}. FedMC~\cite{zou_fedmc_2022} integrates RL models trained by multiple edge devices into a general model based on a meta-learning approach. FedGPO is an RL-based aggregation technique for FL introduced in~\cite{kim_fedgpo_2022}, and aims to optimize the energy-efficiency of FL while guaranteeing model convergence. On the other hand,~\cite{rjoub_trust-driven_2022} introduces DDQN-Trust, a trust-based double deep Q-learning-based selection algorithm for FL that takes into account the trust scores and energy levels of the IoT devices to make appropriate scheduling decisions and integrate it with the main FL aggregation techniques (FedAvg, FedProx, FedShare and FedSGD). Other {federated RL} methods include~\cite{ma_data-aware_2022, tam_privacy-aware_2023, xu_privacy-preserving_2023, peng_online-learning-based_2023, zhang_adaptive_2023, zhao_multi-agent_2023}.

Other {RL} approaches that don't rely on FL or other collaborative learning paradigms have been proposed, such as~\cite{watanabe_fpga-based_2021} which uses online sequential learning to achieve full on-device RL on an FPGA platform. In~\cite{rakesh_dnn_2020}, a method combining supervised and reinforcement learning is proposed for adaptive video streaming on edge servers or on-device, and~\cite{zhang_reinforcement_2022} introduces an on-device RL-based adaptive video transmission algorithm to predict heterogeneous network bandwidth. Finally, RL has also been employed in edge learning via shielding techniques~\cite{DBLP:journals/corr/abs-1708-08611}, as proposed in~\cite{sen_distributed_2022} with a multiagent system that enables each edge node to schedule its own jobs using SROLE, a shielded {RL} technique used to check for action collisions that may occur because of the absence of coordination between the nodes, and provides alternative actions to avoid them.

\subsection{Semi-supervised learning}\label{sec:semi-supervised-learning}
{Semi-supervised learning, is a paradigm that combines labeled and unlabeled data for specific learning tasks~\cite{van2020survey}. It can be described as a middle ground between supervised and unsupervised learning, and leverages the advantages of both approaches. By combining a small amount of labeled with a large amount of unlabeled data. This is particularly beneficial in an edge learning context, where vast amounts of unlabeled data are generated continuously by end devices~\cite{10012038}, and where sending the data for labeling in the cloud is not always possible for privacy reasons. For a comprehensive overview of semi-supervised learning, refer to the following survey~\cite{van2020survey}. Semi-supervised learning presents an alternative mean to harness the vast amount of unlabeled data at the edge. Additionally, it offers other advantage on the edge, as training datasets are often incomplete before training and might need supplementation with} real-time data~\cite{park_semi-supervised_2022}. {Various methodologies have been developed to adapt semi-supervised learning to edge environments, including FL} based approaches~\cite{pei_knowledge_2023, albaseer_exploiting_2020}, {as well as} other learning techniques~\cite{tsukada_neural_2020, zhao_resource-efficient_2022, hou_-device_2020, carta_efficient_2023, wu_onlad-ids_2022, park_semi-supervised_2022, park_semi-supervised_2022, radu_semi-supervised_2014}. 

\subsection{Self-supervised learning}\label{sec:self-supervised-learning}
Self-supervised learning is an ML approach that allows models to learn {from vast amounts of data without explicit labels~\cite{balestriero2023cookbook}. It creates labels from the data itself by defining pretext tasks where the data provides its own supervision. For instance, in natural language processing, a common pretext task involve predicting the context surrounding a word or predicting a given word in a sentence given the previous word, also known as language modeling. In computer vision, a pretext task might involve predicting masked patches of an image. Self-supervised learning} can be especially beneficial in domains where labelled data is scarce, or the specific task can not be known a priori~\cite{balestriero2023cookbook}. {For a deeper exploration on self-supervised learning we refer the reader to~\cite{balestriero2023cookbook}.}

{Because of its independence from labeled data, self-supervised learning is considered as a promising approach for edge learning. By learning from data without explicit labels, it enables learning} useful representations and skills that can be fine-tuned for specific tasks, such as recommendation systems~\cite{xia_efficient_2023, xia_-device_2022}, speech and audio-related applications~\cite{gaol_match_2023, huo_incremental_2022}, and others~\cite{liu_self-train_2022, shi_self-supervised_2023, wu_enabling_2021, wu_federated_2021}. Moreover, self-supervised learning can be particularly effective in scenarios with data and concept drifting~\cite{liu_self-train_2022}. Several recent studies have proposed innovative self-supervised learning methods tailored for edge devices, such as contrastive learning~\cite{liu_self-train_2022, wu_enabling_2021, wu_federated_2021}.

\section{Edge Learning Use Cases and applications}\label{sec:use-case}
As explained in previous sections, {edge learning} offer multiple advantages that range from low latency, bandwidth efficiency to privacy preservation and improved reliability and robustness, it also allows more customization and personalization by adapting to user preferences and behavior without relying on centralized computing or needing to collect and store private user data in cloud servers. In this section, we explore some uses cases and applications for {edge learning} that have been researched and developed in the past years.

\subsection{Healthcare and Remote Monitoring}\label{sec:healthcare}
The use of ML in healthcare has been under constant improvement over the last years~\cite{kitsios_recent_2023}. {However, cloud-based ML solutions still struggle to meet the sector's stringent security requirements~\cite{Qayyum2020SecureAR}, address privacy concerns~\cite{zainuddin_artificial_2021}, and satisfy low latency requirements~\cite{Petersen_2022}. Edge learning has emerged as a promising solution to address these challenges, gaining traction in the field, mostly by using federated edge learning~\cite{das_privacy_2019, qayyum_collaborative_2022, lian_deep-fel_2022, guo_feel_2020}. However, while FL in healthcare is increasingly explored as a privacy-preserving approach, the training of these models is often done with large resource requirements~\cite{chaddadFederatedLearningHealthcare2024, zhouPersonalizedPrivacypreservingFederated2024}, making it hard to implement on edge devices. Moreover, for various tasks, medical data is collected and managed directly by large organizations such as hospitals and medical facilities~\cite{bahriBIGDATAHealthcare2019}. For such tasks FL with more powerful clients with access GPUs may be preferred, since the privacy constraints in these scenarios often involves data not being shared outside the organization rather than the local device. Nevertheless, for tasks where the data shouldn't leave medical edge devices and wearables, such as sensors and ECG devices, edge learning remains a viable and promising direction~\cite{qayyumCollaborativeFederatedLearning2022a, hakakFrameworkEdgeAssistedHealthcare2020}.} Researches that use {edge learning} for healthcare span in most of the field, from atrial fibrillation recognition~\cite{chen_edge2analysis_2022}, preterm labor risk prediction~\cite{chandrika_incremental_2022}, cardiac disease detection~\cite{Wong2023CardiacDisease}, breast ultrasound image classification~\cite{hou_-device_2020} to dermatological disease~\cite{wu_federated_2021} and COVID-19 diagnosis, {leveraging techniques such as CFL~\cite{qayyum_collaborative_2022} and federated transfer learning~\cite{ravi_shanker_reddy_ai_2023}}. {For a more comprehensive overview of FL and edge learning in healthcare, refer to the survey}~\cite{chaddadFederatedLearningHealthcare2024}.

{When it comes to monitoring for medical purposes, Human Activity Recognition (HAR) represents one of the most popular use cases. HAR refers to the automation of the identification and categorization of the various activities performed by humans and their interactions with the environment~\cite{Hari2023HumanActivity}. As personalization for HAR has been shown to improve the results and performance of these systems~\cite{Stojchevska2023EffectivenessPersonalization}, training the model on the edge using incremental learning or meta learning approaches can help achieve that while increasing the privacy and reducing the bandwidth consumption. PILOTE~\cite{zuo_handling_2023} proposes an incremental learning-based approach for HAR, designed for edge devices with extremely limited resources and demonstrates reliable performance in mitigating catastrophic forgetting. In addition,~\cite{lin_model_2020} proposes a personalizable lightweight CNN model for HAR, as well as a training algorithm to find personalization-friendly parameters. With the objective of improving the accuracy after the personalization when dealing with a wide range of target users. ClusterFL~\cite{ClusterFL2023Ouyang}, proposes a clustering-based FL approach for edge-based HAR. Finally,~\cite{craighero_-device_2023} presents an on-device deep learning approach for STM32 microcontrollers, which fine-tunes a CNN model for enhanced HAR personalization.}

\subsection{Smart Technologies}\label{sec:smart_technologies}
Edge learning has emerged as a pivotal technological advancement for smart technologies such as smart cities~\cite{kamruzzaman_new_2021},  smart agriculture~\cite{hayajneh_tiny_2023}, smart homes~\cite{nour_federated_2022}, etc. In this section, we will explore some of its applications in these settings.

Smart cities are urban ecosystems designed using IoT technologies to solve urban life problems and improve the residents' quality of life~\cite{smart_cities_def}. In this context, edge learning has been proposed to solve different challenges, mainly for its ability to leverage ML capabilities while preserving network bandwidth and reducing the charge on cloud servers. In~\cite{na_accelerate_2020}, a cloud-aided edge learning based on knowledge fusion for smart lighting system has been proposed. Another application of ML in the edge is in smart grid systems where ML is needed to improve demand forecasting and automated demand response, as well as to analyze data related to energy use and obtain energy consumption patterns~\cite{smart_grids_ml_need}, detect anomalies~\cite{jithish_distributed_2023}, improve communications~\cite{taik_empowering_2021} and security~\cite{lei_maddpg-based_2021} in the system. Other applications in smart cities include the detection of abnormal and dangerous activities~\cite{huu_detecting_2021, bian_abnormal_2020}, pedestrian detection~\cite{yuan_privacy-preserving_2019}, water consumption forecasting~\cite{el_hanjri_federated_2023, pandiyan_federated_2023} and reducing congestions in intelligent traffic systems~\cite{jaleel_reducing_2020, constantinou_crowd-based_2019}. Other more general {edge learning} approaches for smart cities include~\cite{qolomany_particle_2020, liu_federated_2023, zhang_edge--edge_2019, kamruzzaman_new_2021, albaseer_exploiting_2020}.

Smart farming is another domain where ML is increasingly used to enhance the production quality, crop selection, and mineral deficiency detection, as well as to increase farmers' earnings~\cite{smart_farming_ml_need}. In~\cite{hayajneh_tiny_2023} a TinyML based framework using deep neural networks and LSTM models for {unmanned aerial vehicles} assisted smart farming was proposed, which measure soil moisture and ambient environmental conditions. Smart farming remains a promising domain for edge learning, although further research is needed for effectively harnessing its potential. Finally, using edge learning in smart homes can also be promising, however, at the time of writing this article, only a few papers explore this area, including~\cite{nour_federated_2022}.

\subsection{Autonomous vehicles}\label{sec:auto-vehicles}
Autonomous vehicles are vehicles that can operate without human intervention, they utilize sensor technologies, AI, and networking to navigate and make decisions~\cite{Autonomous_vehicle_explanation}. Autonomous vehicles include self-driving cars, trucks, buses, drones, {Unmanned Aerial Vehicles} (UAVs), and even small robots. As demonstrated in~\cite{yang_lessons_2020} offloading deep learning tasks to edge devices or servers can improve the inference accuracy while meeting the latency constraint, which makes edge learning perfectly suitable for this use-cases, and as expected there has been extensive research done in this area. 

{UAVs} are by far the most prominent use of {edge learning} for autonomous vehicles. In~\cite{noauthor_asynchronous_2023}, a {synchronous FL} structure for multi-UAVs was proposed, that aims to resolve device privacy concerns that come from sending raw data to UAV servers, as well as UAVs' limited processing or communication resources. On the other hand,~\cite{noauthor_model-aided_2023} propose a model-aided federated MARL algorithm to coordinate multiple UAVs on data harvesting missions with limited knowledge about the environment, significantly reducing the real-world training data demand. As mentioned previously in Section~\ref{sec:smart_technologies},~\cite{hayajneh_tiny_2023} aims to assist farming {operations} using UAVs that measure soil moisture and ambient environmental conditions and~\cite{chen_towards_2022} proposes a model to derive computation specifications for learning-based visual odometry from physical characteristics of UAVs. Other {edge learning} applications for UAVs include~\cite{liu_deep_2023, dang_deep-ensemble-learning-based_2022, sharma_feel-enhanced_2023, liu_joint_2023, zhao_predictive_2021, chen_towards_2022, ding_online_2023, tang_battery-constrained_2021, noauthor_opportunistic_2023, cappello_odel_2023}

Edge learning based approaches for other autonomous vehicles are also constantly explored and involve multiple applications, they include:
\begin{itemize}
    \item \textbf{Trajectory predictions} such as~\cite{selvaraj_edge_2021} that proposes a solution for trajectory prediction in the edge for both human-driven and autonomous vehicles by leveraging the capabilities of the {5G multi-access edge computing} platform to collect and process measurements from vehicles and road infrastructure in edge servers and use {an LSTM} model to predict the vehicle trajectory with high accuracy. 
    \item \textbf{Energy efficiency} for autonomous vehicles, where~\cite{zhang_energy-efficient_2022} proposes a rate-splitting multiple access (RSMA)-based Internet of Vehicles system for energy-efficient {FL} in autonomous driving, using {non-orthogonal unicasting and multicasting transmission}.
\end{itemize}

\subsection{Recommendation systems and personalization}\label{sec:recsys}
Recommender systems are intelligent applications that assist users in making decisions by providing advice on products or services they might be interested in~\cite{rec_sys_def}. However, recommender systems that utilize user data can pose threats to user privacy, such as the inadvertent leakage of data to untrusted parties or other users~\cite{recsys_pb_2}. Furthermore, privacy-enhancing techniques may lead to decreased accuracy in the recommendations~\cite{recsys_pb}. Edge learning, and especially collaborative learning approaches such as FL, have a big potential in solving these problems by allowing recommender models to be partially or completely trained on the edge, keeping user interactions on the device and using them to further personalize the system~\cite{yang_federated_2020}.

Different approaches using FL have been used for recommendation systems. Amongst them, FedFast~\cite{muhammad_fedfast_2020} propose to accelerate distributed learning for deep federated recommendation models which achieve high accuracy early in the training process. In~\cite{liu_federated_2022}, a Graph Neural Networks (GNNs) used alongside {FL} for social recommendation tasks, a method that aims to alleviate the cold start problem by inducing information of social links between users~\cite{liu_federated_2022}. On the other hand,~\cite{wei_edge-enabled_2023} was proposed as a federated sequential recommender system for the edge. A method that, unlike traditional recommendations, provides personalized suggestions by sequentially analyzing users' historical interactions~\cite{wang2019sequential}. To achieve this,~\cite{wei_edge-enabled_2023} uses a knowledge-aware transformer and proposed to incorporate knowledge graph information into sequential recommendation tasks, while applying {FL} to preserve users' privacy, and use replaced token detection and two-stream self-attention strategies to enhance the transformer-based model. Finally, FedCT~\cite{liu_fedct_2021} aims to harness  cross-domain recommendation in the edge. While cross-domain recommendation~\cite{zhu2021crossdomain} is a promising area for utilizing data from multiple domains, the conventional approach of sharing data between services in a cloud setting often proves impractical or impossible due to privacy and security concerns. This limitation emphasizes the appeal of {edge learning} as an interesting direction for cross-domain recommendation. By enabling the training of recommender systems on multi-domain data residing on edge devices, while still respecting users' privacy.

Despite the improvement of federated recommender systems for users' privacy preservation~\cite{hu_differentially_2023, guo_prefer_2021}, distributed learning approaches for recommendation still face privacy challenges. Specifically, although users' item ratings remain on-device, they can be inferred from the final model, thereby posing a risk of data leakage when the model is shared with multiple users~\cite{yang_practical_2022, liang_fedrec_2021}. To address this concern, noise is often introduced to the ratings in the form of random user-item interactions. However, this approach usually results in lower performances~\cite{liang_fedrec_2021}. In recent years, several solutions have emerged to mitigate this issue. For instance, FedMMF a federated masked matrix factorization, introduced in~\cite{yang_practical_2022}, aims to protect data privacy in federated recommender systems by using personalized mask generated only from local data. Another approach that aims to achieve that is FedRec++~\cite{liang_fedrec_2021}, by allocating certain clients as denoising clients to eliminate noise {while respecting privacy}, thereby counteracting the random sampling of items during the training phase. Other FL based edge recommendation systems includes~\cite{guo_prefer_2021, du_federated_2021, hu_differentially_2023, dogra_memory_2022, liu_interactive_2022}

Although the most popular approach, FL isn't the only method to train recommender systems in the edge. In~\cite{han_deeprec_2021}, an on-device deep learning sequential recommendation method aimed at mobile devices was proposed, by fine-tuning a pretrained model that was trained using data collected before GDPR\footnote{GDPR: The General Data Protection Regulation is a regulation on data privacy in the European Union and the European Economic Area}, for further personalization. And~\cite{xia_-device_2022} focus on on-device next-item recommendation, and uses compact models and a self-supervised knowledge distillation framework to compensate for the capacity loss caused by compression. Finally,~\cite{qin_split-federated_2023} proposes a split-{FL method} called SpFedRec where a split learning approach was proposed to migrate the item model from participant's edge devices to the cloud side and compress item data while transmitting and apply a Squeeze-and-Excitation network mechanism on the backbone model to optimize the perception of dominant features.

Personalized crowdsourced livecast are another part of personalization methods that might benefit from being {offloaded} to the edge. In~\cite{wang_intelligent_2021} the rapid development of crowdsourced livecast and the challenges in providing personalized {quality of experience} to viewers is discussed, and it introduces an intelligent edge-learning-based framework called ELCast, which integrates {CNNs and deep RL models} in edge computing architectures for personalized crowdcast recommendation. In the area related to video games, personalization involves constructing a system capable of adapting video game rules and content to better suit some aspect of the player preferences, personality, experience and performances~\cite{karpinskyj2014video}. Although not yet explored in edge and on-device learning,~\cite{Bodas2018GamePers} propose a Deep Q network model to personalize games based on user-interaction on the edge.

\subsection{Others}\label{sec:other}
There are multiple other applications and use cases of edge learning that we couldn't explore in this section, from {keyword spotting}~\cite{hard_training_2020, hard_production_2022}, spam detection~\cite{sidhpura_fedspam_2023, sriraman_-device_2022}, IoT threats prediction~\cite{li_predicting_2021}, {camera trap images} classification~\cite{zualkernan_iot_2022}, {detecting defects in photovoltaic components}~\cite{wang_novel_2022}, {estimating air quality}~\cite{chinchole_federated_2021}, to {face spoof attack detection}~\cite{chen_federated_2022} and speech recognition~\cite{soures_enabling_2018, guliani_enabling_2022, guliani_training_2021, sim_robust_2021, park_conformer-based_2023, rao_federated_2023, jia_federated_2022}, another interesting potential application explored in~\cite{pillay_federated_2023} is the use of {edge learning in lunar analogue} environments for future space missions. In general, any use case that benefits from personalization on private user data, or suffers from bandwidth limitations or privacy risks in the training might benefit from fully or partially using edge learning. Therefore, we expect the trend of edge learning to continue rising, expend into other fields and areas and grow beyond the current {use cases} and applications in both academia and the industry.

\section{Libraries, Simulators and Tools for Edge Learning}\label{sec:libs-and-tools}
As an emergent field, {edge learning} requires multiples tools to facilitate its usability, integration and implementation, ranging from emulators and simulators, used train and test ML models on cloud servers before training on the edge, to libraries that allows the successful training of ML models on edge devices.

Although there has been significant work on creating libraries and frameworks for ML at the edge, most of these libraries {focus on} the deployment and inference of {deep learning} on edge devices~\cite{zhang_comprehensive_2023}.{ Only a few libraries enable the training of ML models on edge devices}. {These include} ONNX Runtime\footnote{\label{footnote:onnx-def}\url{https://cloudblogs.microsoft.com/opensource/2023/05/31/on-device-training-efficient-training-on-the-edge-with-onnx-runtime/}}, TensorFlow Lite\footnote{\label{footnote:tf-lite}\url{https://blog.tensorflow.org/2021/11/on-device-training-in-tensorflow-lite.html}} or libraries that focus on distributed learning tasks such as Flower~\cite{beutel_flower_2022} or FedML~\cite{he_fedml_2020}. {Some tools, only allow for researching, prototyping and experimenting of FL methods and are designed for simulating FL methods on the cloud. While others allow for the training and deployment of these techniques in edge devices. Table~\ref{tab:techniques_comparison} shows a list of frameworks intended for edge learning, or for running training simulations for edge learning.}

\begin{table*}[t!]
\caption{{Comparison between The different Frameworks for edge learning}}
\label{tab:framework_comparison}

\resizebox{\textwidth}{!}{

\newcolumntype{C}{>{\centering\arraybackslash}X}
\begin{tabularx}{\fulllength}{p{4cm}CCCCC}

\toprule
\textbf{Framework Name} & \textbf{Support simulation} & \textbf{Allow training on edge device} & \textbf{Type} & \textbf{Language} & \textbf{Plateform} \\

\midrule
{TensorFlow lite} & \xmark & \cmark & Deep Learning & Python  & Android / iOS \\

\midrule
{ONNX Runtime} & \xmark & \cmark & Deep Learning / ML & Multiple Languages  & Android / iOS / Other Plateforms \\

\midrule
{Flower~\cite{beutel_flower_2022}} & \cmark & \cmark & FL & Python  & Android / iOS \\

\midrule
{FedML~\cite{he_fedml_2020}} & \cmark & \cmark & FL & Python  & Android / iOS \\

\midrule
{PySyft~\cite{ziller_pysyft_2021}} & \cmark & \cmark & FL & Python, Kotlin, Swift  & Android / iOS \\

\midrule
{FedERA~\cite{borthakur_federa_2023}}& \cmark & \cmark & FL & Python, Kotlin, Swift  & Android / iOS \\

\midrule
{FedLab~\cite{zeng_fedlab_2023}} & \cmark & \xmark  & FL & Python  & - \\

\midrule
{FedJax~\cite{ro2021fedjax}} & \cmark & \xmark  & FL & Python, Jax  & - \\

\midrule
{LEAF~\cite{caldas_leaf_2019}} & \cmark & \xmark  & FL & Python  & - \\

\midrule
{Flute~\cite{garcia2022flute}} & \cmark & \xmark  & FL & Python  & - \\

\midrule
{PyVertical~\cite{romanini_pyvertical_2021}} & \cmark & \xmark  & FL / Split Learning & Python  & - \\

\midrule
{OpenFL~\cite{foley_openfl_2022}} & \cmark & \xmark  & FL & Python  & - \\

\midrule
{EasyFL~\cite{zhuang_easyfl_2022}} & \cmark & \xmark  & FL & Python  & - \\

\midrule
{FL\_PyTorch~\cite{burlachenko_fl_pytorch_2021}} & \cmark & \xmark  & FL & Python  & - \\

\midrule
{TensorFlow Federated} & \cmark & \xmark  & FL & Python  & - \\

\midrule
{CoreML} & \xmark & \cmark  & ML & Python  & iOS \\

\midrule
{EdgeRL~\cite{park_edgerl_2021}} & \xmark & \cmark  & RL & C/C++  & Embedded Platforms \\

\midrule
{PULP-TrainLib~\cite{nadalini2022pulp}} & \xmark & \cmark  & Deep Learning & C  & RISC-V Multi-core MCUs \\

\bottomrule
        
\end{tabularx}
}

\end{table*}

{PyTorch~\cite{paszke_pytorch_2019} and TensorFlow~\cite{abadi_tensorflow_2016}, the most popular frameworks for training deep learning models, have both developed edge ML libraries}. However, while TensorFlow lite, allows for both the inference and training on the edge, PyTorch mobile\footnote{\label{footnote:pytorch-mobile}\url{https://pytorch.org/mobile/home/}} and ExecuTorch\footnote{\label{footnote:executorch}\url{https://pytorch.org/blog/pytorch-edge/}}, Edge ML libraries for PyTorch at the edge, {only support inference at the time of writing. Note that PyTorch models can be trained using ONNX Runtime\footnote{\label{footnote:onnx-pytorch}\url{https://onnxruntime.ai/blogs/pytorch-on-the-edge}}. ONNX Runtime is a cross-platform ML accelerator with on-device training capabilities. It has deep integration with PyTorch, Hugging Face\footnote{\label{footnote:onnx-hf}\url{https://huggingface.co/blog/optimum-onnxruntime-training}} and TensorFlow, enabling accelerated training and inference on multiple platforms, including mobile devices (Android, iOS) and various hardware accelerators and programming languages. Additionally, ONNX Runtime  supports FL on edge devices through its on-device training capabilities.}

{Over the years, multiple tools and libraries have been proposed to train FL algorithms on edge devices, driven by the need for efficient and decentralized learning. As mentioned earlier, these tools can be categorized into two types. Those that only allow simulation of an edge learning environment in the cloud, and those that allow training on edge devices.}

\paragraph{{Simulation only FL tools}} {Simulation tools are extremely important in the context of edge computing. They are used to model the behavior of fog/edge infrastructures, allowing for the study of interoperability across different layers and protocols in edge-cloud environments~\cite{aral_simulators_2020}. In edge learning, simulators enable experimentation with ML models on cloud servers, facilitating rapid prototyping and experimentation. Many edge learning simulators focus on FL. Notable examples include FL\_PyTorch~\cite{burlachenko_fl_pytorch_2021} a PyTorch-based simulation tool for FL, and TensorFlow Federated\footnote{https://www.tensorflow.org/federated} a similar tool for TensorFlow. Other notable FL simulators are: LEAF~\cite{caldas_leaf_2019}; FedJax~\cite{ro2021fedjax} a JAX-based open source library; Flute~\cite{garcia2022flute} an open source platform with multiple optimization, privacy, and communication strategies; And finally FedLab~\cite{zeng_fedlab_2023} a lightweight open-source framework that focus on algorithm effectiveness and communication efficiency, and allows customization on server optimization, client optimization, communication agreement, and communication compression.}

\paragraph{{FL Simulation tools, which allow the training on an edge device}} {In recent years, numerous libraries have emerged to support the experimentation, development, and deployment of FL algorithms on edge devices. These libraries enable researchers to develop and test FL algorithms on the cloud while facilitating the transition from simulation to real-world deployment on the edge. Notable examples include Flower~\cite{beutel_flower_2022}, and FedML~\cite{he_fedml_2020} which are both aimed toward the research and experimentation of FL algorithms, while allowing the execution of the algorithms on a variety of edge devices. PySyft~\cite{ziller_pysyft_2021} is another FL open-source library that was built as an extension} of PyTorch, Keras, and TensorFlow, and can be run on mobile devices using KotlinSyft\footnote{KotlinSyft: Syft worker for secure on-device machine learning for Android \url{https://github.com/OpenMined/KotlinSyft}} for Android and SwiftSyft\footnote{SwiftSyft: Syft worker for secure on-device machine learning for iOS \url{https://github.com/OpenMined/SwiftSyft}} for iOS. FedERA~\cite{borthakur_federa_2023} {is a similar library that includes a verification module to ensure the validation of local models and avoid aggregating malicious ones. Additionally, FedERA features a carbon emission tracker module to accurately estimate CO2 emissions during the local parameter update phase.}

\paragraph{{Other non-FL Frameworks for edge learning}} {Edge learning frameworks and libraries have been proposed to target specific platforms or hardware. These frameworks aim to simplify the training of ML models on various devices. Notable examples include CoreML, a Swift-based ML inference and training framework for iOS, designed to simplify ML model deployment and training on iOS devices. PULP-TrainLib~\cite{nadalini2022pulp} is another framework, proposed for on-device training on RISC-V multi-core microcontrollers. Additionally, EdgeRL~\cite{park_edgerl_2021} is a lightweight C/C++ framework for on-device reinforcement learning, designed to run on single-core processors typically found in resource-limited embedded platforms.}

\section{Open issues, research directions and future trends}\label{sec:challenges}
In this section, we dive into the challenges, {emerging research paths, and future trends in edge learning. To provide a comprehensive overview, we will divide this section into two parts. The first part will examine the open issues and existing challenges in edge learning, highlighting the obstacles that need to be addressed. The second part will explore promising research directions and our predictions for future trends, shedding light on the opportunities and possibilities that lie ahead.}

\subsection{{Challenges and open issues}}

\subsubsection{{Resource constraints}}
{As highlighted in previous parts of this survey, the resource constraints inherent to edge devices pose the biggest challenge for edge learning. In Section~\ref{sec:overview-techs} we explored the different approaches designed to optimize and accelerate the training of ML models on the edge. However, despite current efforts, limitations in computation, memory, and sometimes energy continue to impede the training of the largest and most complex ML models on the edge. As recent tasks and use cases demand bigger and more complex ML models, the resource limitations of edge devices still pose a significant challenge and remain an open issue. Consequently, ongoing research efforts focus on optimizing ML models for resource constrained environments. Another promising idea, not explored in this survey, involves optimizing edge device hardware for ML~\cite{zhou_-device_2021, marculescu_hardware-aware_2018, dave_hardware_2021}.}

\subsubsection{{Challenges in detecting data quality issues in the edge}}
{Ensuring data quality is a crucial aspect of training ML models~\cite{jain_overview_2020, whang_data_2023}. However, this has proven challenging in the context of edge learning. Due to the decentralized nature of storage inherent in edge computing, detecting data quality issues such as missing or incorrect labels and noisy data is difficult. Additionally, edge devices can be prone to hardware failures, leading to missing or corrupted data~\cite{ergun_dynamic_2023, aral_dependency_2018}. As such, data quality for ML on the edge remains an ongoing challenge~\cite{belgoumri_data_2024}, necessitating further research into developing new methods for detecting and fixing data quality issues, as well as designing ML models that are robust to these issues. The survey~\cite{belgoumri_data_2024} explores the different challenges, constraints, potential solutions, and ongoing efforts related to data quality for ML on the edge.}

\subsubsection{Lack of labelled data availability}
In the context of {edge learning}, a prominent challenge arises from the {prevalence of unlabeled data on edge devices}. This issue becomes particularly problematic as the majority of ML applications traditionally emphasize supervised learning paradigms, necessitating labeled datasets for effective training~\cite{10012038}. {To address this challenge, it is imperative to explore no-label or few-label solutions. This includes a focus on unsupervised (Section~\ref{sec:unsupervised-learning}), self-supervised (Section~\ref{sec:self-supervised-learning}), or semi-supervised (Section~\ref{sec:semi-supervised-learning})} techniques {as well as methods} that can make the most of limited labeled instances such as few-shot learning. Moreover, the development of auto-labeling systems, exemplified by solutions like Flame~\cite{liu_flame_2021}, emerges as a promising {solution to mitigate the impact of the labeled data scarcity on the edge.}

\subsubsection{{Abundance of non-iid data on the edge}}
{As described in Section~\ref{sec:distributed-training-on-edge}, distributed learning methods, such as FL, stand out as pivotal techniques in edge learning. However, a significant body of literature on FL is done under the assumption of IID data~\cite{lu_review_2024}, and very often this assumption doesn't reflect the data present on edge devices. While the effects of Non-IID data on FL depends widely on the type of FL method employed, it often negatively impact the training~\cite{zhu_federated_2021, wang_why_2023, li_convergence_2020}. And while multiple solutions have been proposed to handle Non-IID data, they might come at the expense of privacy preservation and a clear benchmark of real Non-IID performance is unclear considering the large diversity in FL methods~\cite{zhu_federated_2021}. More details on the impact of Non-IID data on FL as well as the specific challenges it poses, and the different approaches proposed to tackle the issue can be found on the following surveys~\cite{lu_review_2024, zhu_federated_2021}.}

\subsubsection{{Data leakage and privacy concerns}}
{Although edge learning aims to provide a privacy-aware alternative to traditional machine learning, it is not immune to privacy and leakage risks during or as a result of the training phase. Notably, model vulnerabilities can be exploited to leak sensitive data, particularly in FL settings where the model is shared among clients~\cite{mothukuriSurveySecurityPrivacy2021, yinComprehensiveSurveyPrivacypreserving2021}. This vulnerability is exacerbated when the model is susceptible to data reconstruction attacks, increasing the risk of private data leakage~\cite{mothukuriSurveySecurityPrivacy2021}. To mitigate these risks, techniques like differential privacy~\cite{weiFederatedLearningDifferential2020, zhang_systematic_2023}, encryption methods~\cite{xieEfficiencyOptimizationTechniques2024}, and other approaches are often employed to ensure optimal privacy. However, further research is necessary to guarantee the preservation of private data and minimize leakage risks in edge learning.}


\subsection{{Future trends and research directions}}

\subsubsection{Hybridization of ML techniques in edge learning}
In the exploration of {edge learning} techniques detailed in Section~\ref{sec:overview-techs}, various strategies have been employed to optimize the training of relatively large ML models on edge devices. However, each of these techniques, as outlined in Table~\ref{tab:techniques_comparison}, comes with distinct advantages and drawbacks. Recognizing the diversity in these methodologies, there has been a notable surge in approaches that advocate for a hybridization of multiple techniques. Figure~\ref{fig:heatmap-techniques} illustrates this promising trend, wherein researchers aim to maximize the advantages and mitigate the drawbacks of individual techniques by combining them. This emerging direction, signifies a deliberate effort to create comprehensive and robust solutions tailored to the unique challenges posed by edge devices. Given the promising results showcased by such hybrid approaches, the trajectory indicates a continued surge in interest and research efforts towards refining and expanding the applicability of hybridized techniques in the domain of edge learning.

\subsubsection{Training large models at the edge}
Large models, including Large Language Models (LLMs)~\cite{touvron2023llama, jiang2024mixtral, team2023gemini}, Diffusion models~\cite{rombach2021highresolution}, and Audio Generation models~\cite{liu2023audioldm, ghosal2023text}, etc., are increasingly prevalent and are steadily growing in popularity. However, despite a growing demand for personalization of these models~\cite{ruiz2023dreambooth, kirk2023personalisation} and privacy concerns in collecting and using personal or private data on centralized cloud servers~\cite{kshetri2023cybercrime}. The fine-tuning, training or personalization of such models at the edge is very challenging considering the limited resources available for edge devices~\cite{woisetschlager2023federated}. Although contributions in this domain are currently limited, the predicted surge in interest prompts a need for proactive exploration. Notable approaches, such as FedLLM\footnote{FedLLM is a platform to Build Large Language Models on Proprietary Data using FL using the FedML Platform \url{https://doc.fedml.ai/federate/fedllm}}, FwdLLM~\cite{xu_federated_2023} and FATE-LLM~\cite{fan2023fate}, have emerged using FL to address the challenges of training LLMs at the edge. Looking ahead, the increasing popularity of LLMs and diffusion models anticipates a growing interest in adapting them for {edge learning}. Furthermore, innovative techniques such as LoRa~\cite{hu2021lora} and Fnet~\cite{lee2021fnet} offer potential solutions for the resource constraints on edge devices, especially when integrated with complementary approaches like FL~\ref{sec:federated-learning}, {split learning}~\ref{sec:split-learning}, or model compression techniques~\ref{sec:compression-training-on-edge}. The convergence of these methodologies holds promise for overcoming challenges associated with training large models at the edge in the foreseeable future.

\subsubsection{Extension to privacy preserving applications}
The escalating popularity of ML applications has brought forth heightened concerns regarding the vast amounts of private and personal data required for effective model training, raising questions about various legal and ethical implications. As discussed in earlier sections, edge learning emerges as a potential solution to address these privacy concerns, as it enables the training of ML models directly on the edge device, eliminating the need for sensitive data to traverse external networks. As such, the usage of edge learning for privacy-preserving applications is expected to be a pivotal research direction for the field. Domains like healthcare (\ref{sec:healthcare}) often had legal requirements as well as ethical concerns of using the data for training ML model~\cite{ramakrishnan2020towards}. And as explored in discussed in previous sections (\ref{sec:recsys}), recommendation systems also stand out as a promising avenue for exploration because the significant scrutiny faced for their reliance on private data during model training~\cite{recsys_pb, recsys_pb_2}. Additionally, other applications that require model personalization and tuning on private data ranging from spam detection in SMS and emails to word suggestions in keyboards and personal assistant chatbots or HAR, can benefit from edge learning, fostering a paradigm shift towards more ethical and privacy-conscious ML applications.

\subsubsection{Reducing energy consumption and carbon footprint}
The {recent surge in distributed learning methodologies has raised concerns} regarding their environmental impact. The substantial energy requirements for training models and data transfer to/from centralized data centers contribute to a significant carbon footprint~\cite{lacoste2019quantifying, castano2023exploring}. Recent trends in ML underscore the critical need to estimate and minimize the environmental impact from the training processes. According to~\cite{PIRSON2021128966}, the estimated carbon footprint associated with edge devices by 2027 will be between 22 and 562 MtCO2-eq/year. Therefore, as a pressing research direction, there is a growing emphasis on developing techniques for {edge learning that consider energy efficiency}. Pioneering works, such as~\cite{savazzi2021framework}, have initiated analyses to quantify the environmental impact of ML in edge devices. Notably, frameworks like FedERA~\cite{borthakur_federa_2023}, designed for training FL models at the edge, incorporate a dedicated carbon emission tracker module to precisely estimate CO2 emissions during the local parameter update phase.

\subsubsection{Frameworks to implement training}
Despite the proliferation of frameworks and libraries aimed at enabling ML on edge devices, the current landscape lacks robust support for on-device training. Existing tools, such as PyTorch Mobile and ExecuTorch, predominantly emphasize inference on mobile/edge devices, neglecting the essential backpropagation algorithms crucial for the training phase. This imbalance in focus between training and inference highlights a critical gap in the current ecosystem. Although some tools have been proposed to facilitate on-device training {on} the edge {(see Section~\ref{sec:libs-and-tools})}, there is a pressing need for the development of new libraries, frameworks, and tools explicitly designed for edge learning.

\section{Conclusion}\label{sec:conclusion}
{This survey aims to provide a comprehensive overview of the vast field of edge learning, which involves training and fine-tuning of ML models at the edge. We have defined edge learning and its associated metrics and requirements, and explored various techniques and methodologies for optimizing ML training at the edge. Additionally, we explored the growing integration} of ML types such as unsupervised learning, reinforcement learning, etc. and the different applications and use cases of {edge learning}. We have also examined the tools, libraries and frameworks used for {edge learning}. {Furthermore, we have} identified key challenges in {edge learning} and attempted to predict future trends and research directions.

{Our analysis has revealed that distributed learning methods, including Federated Learning (FL), are gaining popularity for edge training. We have assessed the benefits and drawbacks of various techniques used to optimize the training at the edge and presented them in Section~\ref{sec:comparison-based-requirements} and Table~\ref{tab:techniques_comparison}. We concluded that distributed techniques such as FL and split learning shows great potential for democratizing edge learning. On the other hand, adaptive and fine-tuning based technique should be considered when possible as they often greatly improve the performances or reduce the training time on the edge significantly. Furthermore, model compression techniques are a great choice when slight decreases in performances are acceptable for a reduced model size and computational requirements. At last, we identified a growing trend towards combining different techniques to mitigate their limitations and maximize their benefits.}

{This survey has provided a broad understanding of edge learning, its requirements, challenges, use cases, and trends, as well as an overview of principal optimization techniques and tools. While it does not provide an in-depth evaluation and comparison of specific task performances, it serves as a reference for developing a foundational understanding of edge learning and identifying areas for future research.}

\section*{Acknowledgments}
This research is supported by the Technology Innovation Institute, UAE under the research contract number TII/DSRC/2022/3143.

\bibliographystyle{IEEEtran}
\bibliography{IEEEabrv, biblio}

\end{document}